%% file: main.tex
\documentclass[lettersize,journal]{IEEEtran}
\IEEEoverridecommandlockouts
\usepackage{amsmath,amsfonts}
\usepackage{algorithmic}
\usepackage[linesnumbered,ruled,vlined]{algorithm2e}
\usepackage{array}
\usepackage[caption=false,font=normalsize,labelfont=sf,textfont=sf]{subfig}
\usepackage{textcomp}
\usepackage{stfloats}
\usepackage{url}
\usepackage{verbatim}
\usepackage{graphicx}
\usepackage{cite}
\usepackage{mhchem}
\usepackage{multirow}
\usepackage[dvipsnames]{xcolor}
\usepackage{lipsum} 
\usepackage{hyperref}
\usepackage{booktabs}
\usepackage{amssymb}
\usepackage{xspace}

\newtheorem{thm}{Theorem}

\newtheorem{proof}{Proof}

\SetKwInput{KwInput}{Input}  
\SetKwInput{KwOutput}{Output}

\makeatletter
\DeclareRobustCommand\onedot{\futurelet\@let@token\@onedot}
\def\@onedot{\ifx\@let@token.\else.\null\fi\xspace}

\def\eg{\emph{e.g}\onedot}

\def\etal{\emph{et al}\onedot}
\makeatother

\newcommand{\mysubsubsection}[1]{\noindent {\bf #1}:}

\begin{document}

\title{Learning to Recover Spectral Reflectance from RGB Images}

\author{Dong Huo, Jian Wang, Yiming Qian and Yee-Hong Yang,~\IEEEmembership{Senior Member,~IEEE}
\thanks{D. Huo, and Y. Yang are with the Department of Computing Science, University of Alberta, Edmonton, AB T6G 2R3, Canada (e-mail:
dhuo@ualberta.ca; herberty@ualberta.ca).

Jian Wang is with Snapchat NYC, New York, NY 10036, USA (e-mail: jwang4@snapchat.com).

Yiming Qian is with the Department of Computer Science, University of
Manitoba, Winnipeg, MB R3T 2N2, Canada (e-mail: qym.ustc@gmail.com).

Digital Object Identifier 10.1109/TIP.2024.3393390

}
}



\maketitle

\input{0_abstract}

\input{1_introduction}

\input{2_review}

\input{3_method}

\input{4_experiments}

\input{5_conclusion}
\input{6_acknowledgment}

\bibliographystyle{IEEEtran}
\bibliography{bib}


\vfill

\end{document}


\title{[Supplementary Document] Learning to Recover Spectral Reflectance from RGB Images}

\author{Dong Huo, Jian Wang, Yiming Qian and Yee-Hong Yang,~\IEEEmembership{Senior Member,~IEEE}


}


\maketitle

This supplementary document provides 1) the detailed derivation of Eqn. 8 in Sec.~\ref{sec:derivation};
2) more qualitative evaluations with competing approaches in Fig.~\ref{fig:supp_qualitative_1}$\sim$\ref{fig:supp_qualitative_3} of Sec.~\ref{sec:more_eva}; 3) recovered reflectance curves and correlation coefficients (corr) between the recovered reflectance and the ground truth in Fig.~\ref{fig:synthetic_curve_1}$\sim$\ref{fig:real_curve_2} of Sec.~\ref{sec:more_eva}; 3) Feasibility analysis of data capture in real world in Sec.~\ref{sec:feas}.

\section{Detailed derivation of Eqn. 8}
\label{sec:derivation}
\begin{align}
\hat{\mathbf{R}}=& (\hat{\omega} + \Delta\omega) (\hat{\boldsymbol{\mathcal{H}}} + \Delta \boldsymbol{\mathcal{H}})^T \cdot ((\hat{\boldsymbol{\mathcal{H}}} + \Delta \boldsymbol{\mathcal{H}}) \cdot (\hat{\boldsymbol{\mathcal{H}}}+ \Delta \boldsymbol{\mathcal{H}})^T)^{-1} \cdot (\boldsymbol{\mathcal{I}} - \Delta\boldsymbol{\mathcal{I}}) + \hat{\mathbf{R}}^\bot \nonumber\\
=& \hat{\omega} (\hat{\boldsymbol{\mathcal{H}}} + \Delta \boldsymbol{\mathcal{H}})^T \cdot ((\hat{\boldsymbol{\mathcal{H}}} + \Delta \boldsymbol{\mathcal{H}}) \cdot (\hat{\boldsymbol{\mathcal{H}}}+ \Delta \boldsymbol{\mathcal{H}})^T)^{-1} \cdot \boldsymbol{\mathcal{I}}
- \hat{\omega} (\hat{\boldsymbol{\mathcal{H}}} + \Delta \boldsymbol{\mathcal{H}})^T \cdot ((\hat{\boldsymbol{\mathcal{H}}} + \Delta \boldsymbol{\mathcal{H}}) \cdot (\hat{\boldsymbol{\mathcal{H}}}+ \Delta \boldsymbol{\mathcal{H}})^T)^{-1} \cdot \Delta\boldsymbol{\mathcal{I}} \nonumber\\
&+\Delta\omega (\hat{\boldsymbol{\mathcal{H}}} + \Delta \boldsymbol{\mathcal{H}})^T \cdot ((\hat{\boldsymbol{\mathcal{H}}} + \Delta \boldsymbol{\mathcal{H}}) \cdot (\hat{\boldsymbol{\mathcal{H}}}+ \Delta \boldsymbol{\mathcal{H}})^T)^{-1} \cdot (\boldsymbol{\mathcal{I}} - \Delta\boldsymbol{\mathcal{I}}) +\hat{\mathbf{R}}^\bot \nonumber\\
=& \hat{\omega} \hat{\boldsymbol{\mathcal{H}}}^T \cdot ((\hat{\boldsymbol{\mathcal{H}}} + \Delta \boldsymbol{\mathcal{H}}) \cdot (\hat{\boldsymbol{\mathcal{H}}}+ \Delta \boldsymbol{\mathcal{H}})^T)^{-1} \cdot \boldsymbol{\mathcal{I}} \nonumber\\
&+ \hat{\omega} \Delta \boldsymbol{\mathcal{H}}^T \cdot ((\hat{\boldsymbol{\mathcal{H}}} + \Delta \boldsymbol{\mathcal{H}}) \cdot (\hat{\boldsymbol{\mathcal{H}}}+ \Delta \boldsymbol{\mathcal{H}})^T)^{-1} \cdot \boldsymbol{\mathcal{I}} 
- \hat{\omega} (\hat{\boldsymbol{\mathcal{H}}} + \Delta \boldsymbol{\mathcal{H}})^T \cdot ((\hat{\boldsymbol{\mathcal{H}}} + \Delta \boldsymbol{\mathcal{H}}) \cdot (\hat{\boldsymbol{\mathcal{H}}}+ \Delta \boldsymbol{\mathcal{H}})^T)^{-1} \cdot \Delta\boldsymbol{\mathcal{I}} \nonumber\\
&+\Delta\omega (\hat{\boldsymbol{\mathcal{H}}} + \Delta \boldsymbol{\mathcal{H}})^T \cdot ((\hat{\boldsymbol{\mathcal{H}}} + \Delta \boldsymbol{\mathcal{H}}) \cdot (\hat{\boldsymbol{\mathcal{H}}}+ \Delta \boldsymbol{\mathcal{H}})^T)^{-1} \cdot (\boldsymbol{\mathcal{I}} - \Delta\boldsymbol{\mathcal{I}}) +\hat{\mathbf{R}}^\bot \nonumber\\
= &\hat{\omega} \hat{\boldsymbol{\mathcal{H}}}^T \cdot (\hat{\boldsymbol{\mathcal{H}}} \cdot \hat{\boldsymbol{\mathcal{H}}}^T + \hat{\boldsymbol{\mathcal{H}}} \cdot \Delta\boldsymbol{\mathcal{H}}^T + \Delta\boldsymbol{\mathcal{H}} \cdot \hat{\boldsymbol{\mathcal{H}}}^T + \Delta\boldsymbol{\mathcal{H}} \cdot \Delta\boldsymbol{\mathcal{H}}^T)^{-1} \cdot \boldsymbol{\mathcal{I}} \nonumber\\
&+ \hat{\omega}\Delta \boldsymbol{\mathcal{H}}^T \cdot ((\hat{\boldsymbol{\mathcal{H}}} + \Delta \boldsymbol{\mathcal{H}}) \cdot (\hat{\boldsymbol{\mathcal{H}}}+ \Delta \boldsymbol{\mathcal{H}})^T)^{-1} \cdot \boldsymbol{\mathcal{I}}
- \hat{\omega} (\hat{\boldsymbol{\mathcal{H}}} + \Delta \boldsymbol{\mathcal{H}})^T \cdot ((\hat{\boldsymbol{\mathcal{H}}} + \Delta \boldsymbol{\mathcal{H}}) \cdot (\hat{\boldsymbol{\mathcal{H}}}+ \Delta \boldsymbol{\mathcal{H}})^T)^{-1} \cdot \Delta\boldsymbol{\mathcal{I}}\nonumber\\
&+\Delta\omega (\hat{\boldsymbol{\mathcal{H}}} + \Delta \boldsymbol{\mathcal{H}})^T \cdot ((\hat{\boldsymbol{\mathcal{H}}} + \Delta \boldsymbol{\mathcal{H}}) \cdot (\hat{\boldsymbol{\mathcal{H}}}+ \Delta \boldsymbol{\mathcal{H}})^T)^{-1} \cdot (\boldsymbol{\mathcal{I}} - \Delta\boldsymbol{\mathcal{I}}) +\hat{\mathbf{R}}^\bot.\tag{17}
\label{eqn:detail_1}
\end{align}
Define the singular value decomposition of $\hat{\boldsymbol{\mathcal{H}}} \cdot \Delta\boldsymbol{\mathcal{H}}^T + \Delta\boldsymbol{\mathcal{H}} \cdot \hat{\boldsymbol{\mathcal{H}}}^T + \Delta\boldsymbol{\mathcal{H}} \cdot \Delta\boldsymbol{\mathcal{H}}^T$ as 
\begin{equation}
\boldsymbol{U} \cdot \boldsymbol{\Sigma} \cdot \boldsymbol{V}^T = SVD (\hat{\boldsymbol{\mathcal{H}}} \cdot \Delta\boldsymbol{\mathcal{H}}^T + \Delta\boldsymbol{\mathcal{H}} \cdot \hat{\boldsymbol{\mathcal{H}}}^T + \Delta\boldsymbol{\mathcal{H}} \cdot \Delta\boldsymbol{\mathcal{H}}^T).\tag{18}
\label{eqn:svd}
\end{equation}
Following the derivation of Henderson and Searle~\cite{henderson1981deriving}, Eqn.~\ref{eqn:detail_1} can be reformulated as
\begin{align}
\hat{\mathbf{R}}
= &\hat{\omega} \hat{\boldsymbol{\mathcal{H}}}^T \cdot 
(
(\hat{\boldsymbol{\mathcal{H}}} \cdot \hat{\boldsymbol{\mathcal{H}}}^T)^{-1} 
-
(\hat{\boldsymbol{\mathcal{H}}} \cdot \hat{\boldsymbol{\mathcal{H}}}^T)^{-1} \cdot \boldsymbol{U} \cdot 
(\boldsymbol{I} + \boldsymbol{\Sigma} \cdot \boldsymbol{V}^T \cdot (\hat{\boldsymbol{\mathcal{H}}} \cdot \hat{\boldsymbol{\mathcal{H}}}^T)^{-1} \cdot \boldsymbol{U})^{-1} 
\cdot \boldsymbol{\Sigma} \cdot \boldsymbol{V}^T \cdot (\hat{\boldsymbol{\mathcal{H}}} \cdot \hat{\boldsymbol{\mathcal{H}}}^T)^{-1}) \cdot \boldsymbol{\mathcal{I}} \nonumber\\
&+ \hat{\omega}\Delta \boldsymbol{\mathcal{H}}^T \cdot ((\hat{\boldsymbol{\mathcal{H}}} + \Delta \boldsymbol{\mathcal{H}}) \cdot (\hat{\boldsymbol{\mathcal{H}}}+ \Delta \boldsymbol{\mathcal{H}})^T)^{-1} \cdot \boldsymbol{\mathcal{I}}
- \hat{\omega} (\hat{\boldsymbol{\mathcal{H}}} + \Delta \boldsymbol{\mathcal{H}})^T \cdot ((\hat{\boldsymbol{\mathcal{H}}} + \Delta \boldsymbol{\mathcal{H}}) \cdot (\hat{\boldsymbol{\mathcal{H}}}+ \Delta \boldsymbol{\mathcal{H}})^T)^{-1} \cdot \Delta\boldsymbol{\mathcal{I}}\nonumber\\
&+\Delta\omega (\hat{\boldsymbol{\mathcal{H}}} + \Delta \boldsymbol{\mathcal{H}})^T \cdot ((\hat{\boldsymbol{\mathcal{H}}} + \Delta \boldsymbol{\mathcal{H}}) \cdot (\hat{\boldsymbol{\mathcal{H}}}+ \Delta \boldsymbol{\mathcal{H}})^T)^{-1} \cdot (\boldsymbol{\mathcal{I}} - \Delta\boldsymbol{\mathcal{I}}) +\hat{\mathbf{R}}^\bot\nonumber\\
= &\hat{\omega} \hat{\boldsymbol{\mathcal{H}}}^T \cdot (\hat{\boldsymbol{\mathcal{H}}} \cdot \hat{\boldsymbol{\mathcal{H}}}^T)^{-1} \cdot \boldsymbol{\mathcal{I}} \nonumber\\
&- \hat{\omega} \hat{\boldsymbol{\mathcal{H}}}^T \cdot((\hat{\boldsymbol{\mathcal{H}}} \cdot \hat{\boldsymbol{\mathcal{H}}}^T)^{-1} \cdot \boldsymbol{U} \cdot 
(\boldsymbol{I} + \boldsymbol{\Sigma} \cdot \boldsymbol{V}^T \cdot (\hat{\boldsymbol{\mathcal{H}}} \cdot \hat{\boldsymbol{\mathcal{H}}}^T)^{-1} \cdot \boldsymbol{U})^{-1} 
\cdot \boldsymbol{\Sigma} \cdot \boldsymbol{V}^T \cdot (\hat{\boldsymbol{\mathcal{H}}} \cdot \hat{\boldsymbol{\mathcal{H}}}^T)^{-1}) \cdot \boldsymbol{\mathcal{I}} \nonumber\\
&+ \hat{\omega}\Delta \boldsymbol{\mathcal{H}}^T \cdot ((\hat{\boldsymbol{\mathcal{H}}} + \Delta \boldsymbol{\mathcal{H}}) \cdot (\hat{\boldsymbol{\mathcal{H}}}+ \Delta \boldsymbol{\mathcal{H}})^T)^{-1} \cdot \boldsymbol{\mathcal{I}}
- \hat{\omega} (\hat{\boldsymbol{\mathcal{H}}} + \Delta \boldsymbol{\mathcal{H}})^T \cdot ((\hat{\boldsymbol{\mathcal{H}}} + \Delta \boldsymbol{\mathcal{H}}) \cdot (\hat{\boldsymbol{\mathcal{H}}}+ \Delta \boldsymbol{\mathcal{H}})^T)^{-1} \cdot \Delta\boldsymbol{\mathcal{I}}\nonumber\\
&+\Delta\omega (\hat{\boldsymbol{\mathcal{H}}} + \Delta \boldsymbol{\mathcal{H}})^T \cdot ((\hat{\boldsymbol{\mathcal{H}}} + \Delta \boldsymbol{\mathcal{H}}) \cdot (\hat{\boldsymbol{\mathcal{H}}}+ \Delta \boldsymbol{\mathcal{H}})^T)^{-1} \cdot (\boldsymbol{\mathcal{I}} - \Delta\boldsymbol{\mathcal{I}}) +\hat{\mathbf{R}}^\bot\nonumber\\
=& \hat{\omega} \underbrace{\hat{\boldsymbol{\mathcal{H}}}^T \cdot (\hat{\boldsymbol{\mathcal{H}}} \cdot \hat{\boldsymbol{\mathcal{H}}}^T)^{-1} \cdot \boldsymbol{\mathcal{I}}}_{\hat{\mathbf{R}}_{\hat{\boldsymbol{\mathcal{H}}}}} + \Delta \hat{\mathbf{R}},\tag{19}
\label{eqn:detail_2}
\end{align}
where $\mathcal{I}$ represents the identity matrix.

\section{More Evaluation Results}
\label{sec:more_eva}

\input{figures/supp_qualitative_1}
\input{figures/supp_qualitative_2}
\input{figures/supp_qualitative_3}

\input{figures/synthetic_curve_1.tex}
\input{figures/synthetic_curve_2.tex}
\input{figures/real_curve_1.tex}
\input{figures/real_curve_2.tex}


\clearpage
\newpage

\input{figures/iphone}
\section{Feasibility analysis of data capture}
\label{sec:feas}

Our method needs RGB images of the same scene under different illuminations as the input. Capturing RGB images of the same scene under different illuminations is feasible and has been realized in tasks like photometric stereo. To our knowledge, two options exist. Firstly, sequential acquisition is common if the scene is static. Secondly, a commodity high-speed camera such as iPhone can be utilized. Specifically, consider the exposure time of RGB images is short, one could record high-speed videos (120/240 FPS) with alternating light sources to obtain images under different illuminations as in~\cite{park2007multispectral}. The iPhone flashlights consist of both white and amber LEDs (see Fig.~\ref{fig:iphone}), which can be used as alternating light sources. The impact of ambient light can be removed by subtracting images captured with white/amber LEDs off. Small motion in high-speed videos is negligible. Extremely fast object/camera motion is beyond the scope of this paper. In addition to exploit the flashlight and the rear-facing camera of an iPhone, one could also explore the screen light and the front-facing camera.


\bibliographystyle{IEEEtran}
\bibliography{bib}

\vfill

%% file: 0_abstract.tex
\begin{abstract}



This paper tackles spectral reflectance recovery (SRR) from RGB images. Since capturing ground-truth spectral reflectance and camera spectral sensitivity are challenging and costly, most existing approaches are trained on synthetic images and utilize the same parameters for all unseen testing images, which are suboptimal especially when the trained models are tested on real images because they never exploit the internal information of the testing images. To address this issue, we adopt a self-supervised meta-auxiliary learning (MAXL) strategy that fine-tunes the well-trained network parameters with each testing image to combine external with internal information. To the best of our knowledge, this is the first work that successfully adapts the MAXL strategy to this problem. Instead of relying on naive end-to-end training, we also propose a novel architecture that integrates the physical relationship between the spectral reflectance and the corresponding RGB images into the network based on our mathematical analysis. Besides, since the spectral reflectance of a scene is independent to its illumination while the corresponding RGB images are not, we recover the spectral reflectance of a scene from its RGB images  captured under multiple illuminations to further reduce the unknown. Qualitative and quantitative evaluations demonstrate the effectiveness of our proposed network and of the MAXL. Our code and data are available at \href{https://github.com/Dong-Huo/SRR-MAXL}{https://github.com/Dong-Huo/SRR-MAXL}.
\end{abstract}
\begin{IEEEkeywords}
Spectral reflectance recovery, multiple illuminations, sub-space components, meta-auxiliary learning
\end{IEEEkeywords}

%% file: 1_introduction.tex
\section{Introduction}
\label{sec:intro}
Unlike traditional RGB images with only three bands (red, green, and blue), the spectral reflectance captured by a hyperspectral imaging system has a higher sampling rate in wavelength and provides more spectral information of the scene. The spectral reflectance of an object is independent of the illumination so that it describes the distinctive intrinsic characteristics of an object's materials, which is widely used in many applications such as remote sensing~\cite{goetz1985imaging, nanni2006spectral}, agriculture~\cite{carter1993responses, martinez2005multispectral}, medical imaging~\cite{tsumura2001medical, preece2002monte, kim2016smartphone, he2020hyperspectral}, and food quality evaluation~\cite{wang2016recent, elmasry2012meat}.



Despite certain snapshot hyperspectral imaging systems~\cite{hagen2013review} capable of capturing spectral reflectance at  high frame rates, their performance is limited by low spatial resolution and low spectral accuracy. Consequently, the acquisition of precise and high-quality spectral reflectance remains a time-consuming process. Hyperspectral imaging systems capture one or two dimensions of the three-dimensional datacube at a time, and sequentially scan (area scanning, point scanning, or line scanning) the remaining dimension(s) for the complete datacube~\cite{elmasry2010principles}, which are not suitable for dynamic scenes. An alternative approach is to recover the spectral reflectance from RGB images~\cite{cao2017spectral, wu2019reflectance, deeb2018spectral, fu2018spectral, nguyen2014training, park2007multispectral}. 


\input{figures/teaser}

Ever since the emergence of deep neural networks (DNNs), spectral reconstruction from RGB images has achieved impressive results using end-to-end training~\cite{sun2021tuning, cai2022mst++, alvarez2017adversarial, zhang2020pixel, fu2018joint, shi2018hscnn+, yan2018accurate}. Spectral reconstruction can be categorised into two main classes: hyperspectral image reconstruction (HIR) and spectral reflectance recovery (SRR), where a hyperspetral image is factorized as a product of a spectral reflectance and an illumination spectrum. In this paper, we focus on the SRR with known illuminations. 

The main shortcoming of existing DNN-based methods is that they apply the same trained parameters to all testing images and fail to utilize image-specific information, resulting in sub-optimal solutions~\cite{park2020fast} because of the domain shift, especially when they are trained on synthetic data but tested on real data. One possible solution to overcome this issue is utilizing zero-shot learning~\cite{shocher2018zero, ren2020neural, huo2022blind, rasti2021undip, sidorov2019deep} that directly extracts the internal information of a given testing image in a self-supervised manner. However, the limited information on a single image may not be enough to optimize the network, and the optimization time for each testing image is long (usually takes minutes or even hours). Besides, existing DNN-based methods rely on end-to-end training which do not take the physical properties of the spectral reflectance into consideration.

This paper takes a step forward using meta-auxiliary learning (MAXL)~\cite{chi2021test} to take advantage of both internal and external information for SRR under known illuminations, with the goal to rapidly adapt the trained parameters to an unseen image using only a few steps of gradient descent at test time. In particular, we design a neural architecture featuring two tasks: the primary task focuses on recovering spectral reflectance from RGB images, while the auxiliary task involves reconstructing RGB inputs from the recovered spectral reflectance.  We adopt both tasks to train the model on paired inputs and outputs (referred to as external information), and fine-tune the pretrained parameters using a single testing input (referred to as internal information) leveraging solely the auxiliary task. Notably, the fine-tuning process eliminates the need for paired ground truth. Experiments show that MAXL significantly boosts the performance on real data, which demonstrates the effectiveness of MAXL in reducing domain gap.

In addition, following our mathematical analysis in Section~\ref{sec:formulation}, we propose a novel architecture that explicitly utilizes the sub-space of a camera spectral sensitivity (CSS) and illumination spectra to integrate the physical relationship between RGB images and the corresponding spectral reflectances into the network, instead of relying on naive end-to-end training. Lin~\etal~\cite{lin2020physically} also attempt to leverage the sub-space of a CSS for HIR. They assume that a spectrum is the summation of components in the sub-space and the null-space of a known CSS. Since sub-space components can be directly obtained from the CSS and the RGB image, they directly estimate the null-space components using end-to-end training. However, they have not considered the information loss when discretizing a continuous spectrum for RGB synthesis, so that the assumption is no longer satisfied on real data. In contrast, we design our network to compensate for the information loss of discretization under a varying number of illuminations of a scene captured with a camera with an unknown CSS which varies from device to device.

For the illumination, we adopt white and amber LEDs which are ubiquitous on mobile devices (as shown in Fig.~\ref{fig:teaser}), instead of complicated multiplexed illuminations~\cite{park2007multispectral, tschannerl2019hyperspectral, goel2015hypercam}. 

Our contributions are summarized below:
\begin{itemize}
\item We propose a novel architecture motivated by our mathematical derivation that integrates physical properties of the spectral reflectance into the network with an unknown camera spectral sensitivity (CSS);
\item We propose a unified framework for recovering spectral reflectance from RGB images captured under more than one illumination;
\item We present the first work that successfully adopt meta-auxilary learning (MAXL) to spectral reflectance recovery (SRR). To the best of our knowledge, it is the first attempt to explore the potential of MAXL in this task.
\item Our proposed method dramatically outperforms state-of-the-art methods on both synthetic data and our collected real data.

\end{itemize}



%% file: figures/teaser.tex
\begin{figure}
\centering
\includegraphics[width=0.47\textwidth]{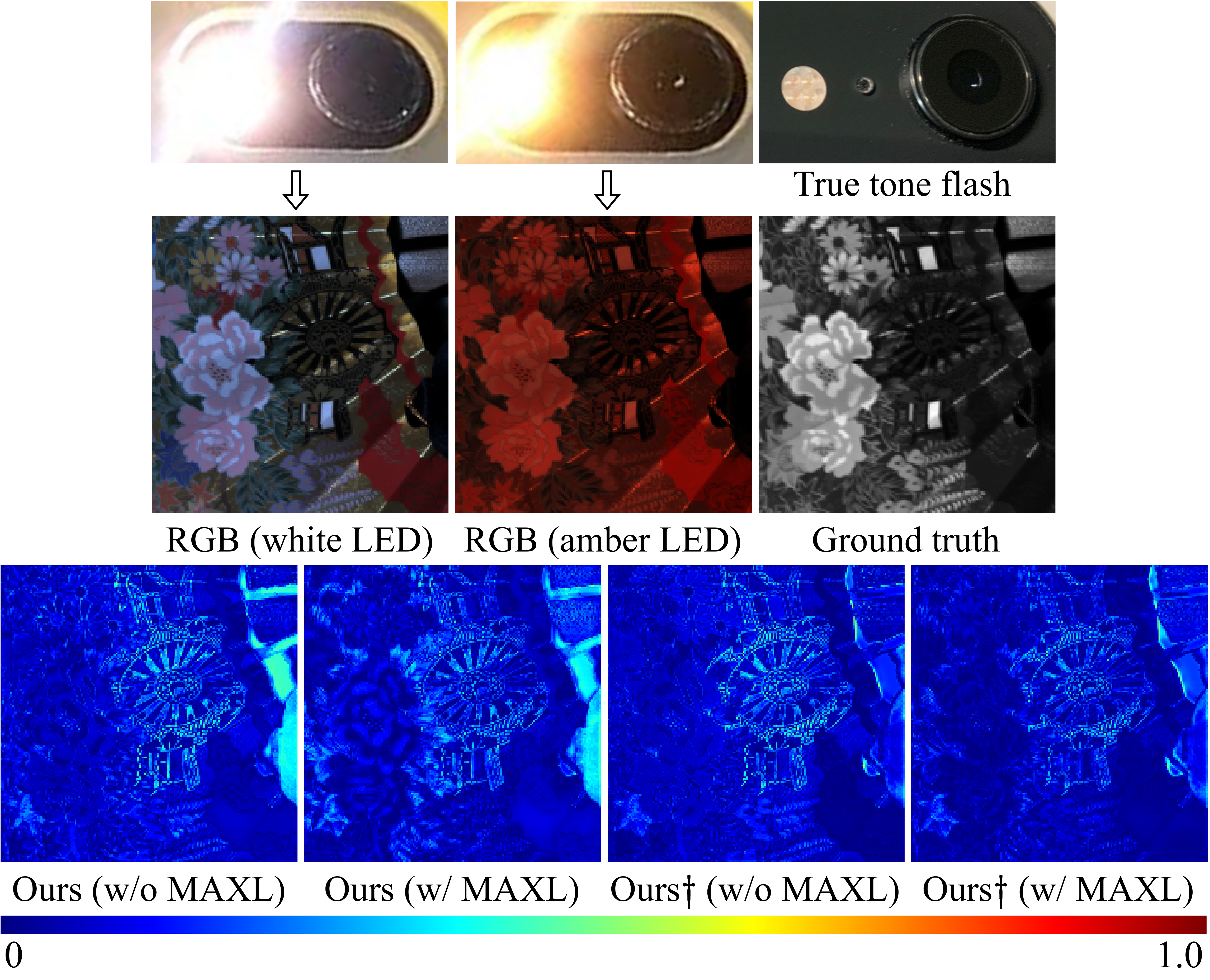}
\caption{This paper proposes a novel spectral reflectance recovery approach from RGB images, which utilizes meta-auxiliary learning (MAXL) to exploit the internal information from testing images. It also demonstrates that an extra illumination (amber LED) can benefit the performance compared with a single illumination (white LED). The illumination LEDs could come from the True Tone flash \cite{pcmag_truetone} of smartphones (first row).
The last row shows the error maps of the recovered results, where Ours and Ours$\dagger$ represent the model w/o and w/ the extra illumination, respectively. }
\label{fig:teaser}
\vspace{-10pt}
\end{figure}

%% file: 2_review.tex
\section{Related Work}
\subsection{Spectral Reconstruction from RGB}

\mysubsubsection{Conventional methods} The spectral reflectance of a scene can be represented by a linear combination of several base spectra~\cite{chakrabarti2011statistics}. Conventional methods mainly focus on learning the base spectra and the corresponding representation coefficients~\cite{fu2018spectral, arad2016sparse, akhtar2018hyperspectral, jia2017rgb, timofte2015a+}. For example, Arad and Ben-Shahar~\cite{arad2016sparse} create an over complete hyperspectral dictionary using K-SVD and learn the representation coefficients from the RGB counterpart. Fu~\etal~\cite{fu2018spectral} first cluster the hyperspectral data and create a dictionary for each cluster, and the spectral reflectance of each pixel is learned from its nearest cluster. Jia~\etal~\cite{jia2017rgb} utilize a low-dimensional manifold to represent the high-dimensional spectral data, which is able to learn a well-conditioned three-to-three mapping between a RGB vector and a 3D point in the embedded natural spectra. Akhtar and Mian~\cite{akhtar2018hyperspectral} also cluster the spectral data but replace the dictionary with Gaussian processes.


\mysubsubsection{DNN-based methods} Recently, DNN-based methods have dominated this area owing to the encouraging results of external learning~\cite{sun2021tuning, shi2018hscnn+, cai2022mst++, zhang2020pixel, xiong2017hscnn, fu2018joint, zhao2020hierarchical, yan2018accurate, stiebel2018reconstructing, li2022drcr, lin2020physically}. Shi~\etal~\cite{shi2018hscnn+} stack multiple residual blocks or dense blocks for end-to-end spectral reconstruction. Lin~\etal~\cite{lin2020physically} separate the spectra into the sub-space and the null-space of the CSS for plausible reconstruction, where the sub-space component signifies the projection of the spectra onto the CSS matrix, while the null-space component represents the remaining portion. Our approach builds upon this concept by extending it to the recovery of spectral reflectance in cases where the CSS is unknown. Zhang~\etal~\cite{zhang2020pixel} generate basis functions from different receptive fields and fuse them with learned pixel-wise weights. Sun~\etal~\cite{sun2021tuning} estimate the spectral reflectance and the illumination spectrum simultaneously with a learnable IR-cut filter. Hang~\etal~\cite{hang2021spectral} decompose the spectral bands into groups based on the correlation coefficients and estimate each group separately using a neural network. A self-supervised loss further constrains the reconstruction. Li~\etal~\cite{li2022drcr} exploit channel-wise attention to refine the degraded RGB images. Cai~\etal~\cite{cai2022mst++} exploit the spectral-wise self-attention to capture inter-spectra correlations.  Li~\etal~\cite{li2022quantization} learn a quantized diffractive optical element (DOE) to improve the hyperspectral imaging of RGB cameras. Zhang~\etal~\cite{zhang2022implicit} exploit the implicit neural representation that maps a spatial coordinate to the corresponding continuous spectrum using a multi-layer perceptron (MLP) whose parameters are generated from a convolution network. Some methods guide the reconstruction with a low-resolution hyperspectral image~\cite{dong2021model, wang2019deep, zhu2020hyperspectral}, which are different from the scope of this paper. All of the above mentioned methods do not considered the internal information from testing cases.

\subsection{Meta-auxiliary Learning}
In contrast with the term “meta-auxiliary learning (MAXL)" in image classification~\cite{liu2019self}, which is designed to improve the generalization of classification models by using meta-learning~\cite{andrychowicz2016learning} to discover optimal labels for auxiliary tasks without the need of manually-labelled auxiliary data~\cite{zhang2018fine}, Chi~\etal~\cite{chi2021test} redefine MAXL as a combination of model-agnostic meta-learning (MAML)~\cite{finn2017model} and auxiliary-learning (AL)~\cite{sun2020test} for test-time fast adaptation. In this paper, we use the definition of the latter.

\mysubsubsection{Model-agnostic meta-learning (MAML)} The aim of MAML is to train models capable of fast adaptation to a new task with only a few steps of gradient descent, which can be applied to few-shot learning~\cite{sun2019meta}. Park~\etal~\cite{park2020fast} and Soh~\etal~\cite{soh2020meta} adopt the MAML for super-resolution. They first initialize the model by training on external datasets like other DNN-based super resolution methods~\cite{wang2020deep}, then conduct MAML to further optimize the model for unseen kernels. During testing, a low resolution input and its down-scaled version are represented as a new training pair to fine-tune the model. Although the targets are similar (spatial/spectral upsampling), directly applying the MAML to SRR is infeasible because the three RGB channels of an image cannot be further downsampled.

\mysubsubsection{Auxiliary-learning (AL)} AL is to assist the optimization of primary tasks with at least one auxiliary task for better generalization and performance. Guo~\etal~\cite{guo2020closed} reconstruct low resolution images for real-world super resolution. Valada~\etal~\cite{valada2018deep} learn to estimate visual odometry and global pose simultaneously for higher efficiency. Lu~\etal~\cite{lu2020depth} solve the depth completion problem with image reconstruction to extract more semantic cues. AL can also stabilize the training of GAN for image synthesis~\cite{odena2017conditional}. Sun~\etal~\cite{sun2020test} choose the rotation prediction as the auxiliary task to update pre-trained parameters for test-time adaptation. Nevertheless, simply updating the pre-trained parameters with only auxiliary tasks may result in catastrophic forgetting~\cite{chi2021test}, where the model exhibits overfitting in the auxiliary tasks, leading to a loss of acquired knowledge from the primary task during the training process.


To leverage both the MAML and AL, we follow the strategy of Chi~\etal~\cite{chi2021test} using self-supervised RGB reconstruction as the auxiliary task and MAML to avoid catastrophic forgetting. The auxiliary task also avoids downsampling RGB images to generate training pairs for fine-tuning.

%% file: 3_method.tex
\section{Method}
\subsection{Problem Formulation}
\label{sec:formulation}
The relationship between an RGB image of a scene and its spectral reflectance can be expressed as
\begin{equation}
I^c (x, y) = \int_\lambda S^c (\lambda) L (\lambda) R (x, y, \lambda) d\lambda,
\label{eqn:integral}
\end{equation}
where $I^c$ represents channel $c$ of the RGB image ($c\in \{Red, Green, Blue\}$), $R$ the spectral reflectance, $S^c$ the CSS of channel $c$, and $L$ the illumination spectrum. $\lambda$ refers to the wavelength, and $(x, y)$ are the spatial coordinates. Assume that the number of pixels and the number of sampled spectral bands are $N$ and $B$, respectively, Eqn.~\ref{eqn:integral} can be discretized and represented in matrix form as
\begin{equation}
\mathbf{I} = (\mathbf{S} \odot \mathbf{L}) \cdot \mathbf{R},
\label{eqn:mm}
\end{equation}
where $\mathbf{I}\in \mathbb{R}^{3\times N}$ is the RGB image, $\mathbf{R}\in \mathbb{R}^{B\times N}$ is the spectral reflectance, $\mathbf{S}\in \mathbb{R}^{3\times B}$ denotes the CSS, and $\mathbf{L}\in \mathbb{R}^{1\times B}$ denotes the illumination spectrum. $\odot$ is the Hadamard product, and $\cdot$ is the matrix multiplication. 

Since the system is under-determined, more images of the same scene under different and independent $\mathbf{L}$ can help to reduce the unknown, which can be formulated as
\begin{equation}
\boldsymbol{\mathcal{I}} = \boldsymbol{\mathcal{H}} \cdot \mathbf{R},\;
\boldsymbol{\mathcal{I}} = 
\begin{bmatrix}
         \mathbf{I}_1\\
         \vdots\\ 
         \mathbf{I}_M
\end{bmatrix},\;
\boldsymbol{\mathcal{H}} =
\begin{bmatrix}
         \mathbf{S} \odot \mathbf{L}_1\\
         \vdots\\ 
         \mathbf{S} \odot \mathbf{L}_M\\ 
\end{bmatrix},
\label{eqn:more_mm}
\end{equation}
where $M$ is the number of illuminations, $\boldsymbol{\mathcal{I}}\in \mathbb{R}^{3M\times N}$ is the stack of RGB images of the same scene. Our goal is to learn a mapping $F(\cdot)$ from $\boldsymbol{\mathcal{I}}$ to $\mathbf{R}$ with known illuminations and unknown CSSs, as
\begin{equation}
\hat{\mathbf{R}} = F (\boldsymbol{\mathcal{I}}, \mathbf{L}_1, \hdots, \mathbf{L}_M).
\label{eqn:primary}
\end{equation}
Instead of naively learning an end-to-end mapping between $\boldsymbol{\mathcal{I}}$ and $\mathbf{R}$, we attempt to take $\boldsymbol{\mathcal{H}}$ into consideration so that the physical relationship of $\boldsymbol{\mathcal{I}}$ and $\mathbf{R}$ can be exploited. 

Lin~\etal~\cite{lin2020physically} prove that all possible solutions of $\hat{\mathbf{R}}$ shares the same component $\hat{\mathbf{R}}^\parallel$ within the sub-space of $\boldsymbol{\mathcal{H}}$, where
\begin{equation}
\hat{\mathbf{R}}^\parallel = \boldsymbol{\mathcal{H}}^T \cdot (\boldsymbol{\mathcal{H}} \cdot \boldsymbol{\mathcal{H}}^T)^{-1} \cdot \boldsymbol{\mathcal{I}}.
\label{eqn:r_parallel}
\end{equation}
As we can see, $\hat{\mathbf{R}}^\parallel$ can be directly calculated from $\boldsymbol{\mathcal{H}}$ and $\boldsymbol{\mathcal{I}}$, so that they aim at learning the other component $\hat{\mathbf{R}}^\bot$ within the null-space of $\boldsymbol{\mathcal{H}}$, and the recovered result is $\hat{\mathbf{R}}^\parallel+\hat{\mathbf{R}}^\bot$. Nevertheless, simply adopting this strategy to our problem may lead to the following issues:

\input{figures/spectrum}

\begin{itemize}
\item Real RGB images are integrated from continuous spectra as in Eqn.~\ref{eqn:integral}, the discretized form in Eqn.~\ref{eqn:mm} and~\ref{eqn:more_mm} are obtained by sub-sampling, resulting in information loss on RGB images (as shown in Fig.~\ref{fig:spectrum});
\item $\mathbf{S}$ is unknown because it varies from sensor to sensor. We have to train an extra network to estimate it from $\boldsymbol{\mathcal{I}}$ and approximate the matrix $\boldsymbol{\mathcal{H}}$;
\item The real intensity of illumination depends on the standard exposure settings~\cite{lin2020physically}, but our illumination spectra are normalized to $[0, 1]$ which need to be rescaled with factor $\omega$;
\item We empirically observe that the back-propagation of the null-space is extremely unstable.
\end{itemize}


\input{figures/architecture}

To solve the above mentioned problems, the recovered result needs to be reformulated as
\begin{equation}
\hat{\mathbf{R}} = \hat{\omega} \hat{\mathbf{R}}_{\hat{\boldsymbol{\mathcal{H}}}} + \Delta \hat{\mathbf{R}},
\label{eqn:r_parallel_redefine}
\end{equation}
\begin{equation}
\hat{\mathbf{R}}_{\hat{\boldsymbol{\mathcal{H}}}} = \hat{\boldsymbol{\mathcal{H}}}^T \cdot (\hat{\boldsymbol{\mathcal{H}}} \cdot \hat{\boldsymbol{\mathcal{H}}}^T)^{-1} \cdot \boldsymbol{\mathcal{I}},\;
\hat{\boldsymbol{\mathcal{H}}} =
\begin{bmatrix}
         \hat{\mathbf{S}} \odot \mathbf{L}_1\\
         \vdots\\ 
         \hat{\mathbf{S}} \odot \mathbf{L}_M\\ 
\end{bmatrix},
\label{eqn:rh_hat}
\end{equation}
where $\hat{\omega}$ is the estimated rescaling factor for the illumination, and $\hat{\mathbf{S}}$ denotes the estimated CSS. We directly generate $\Delta \hat{\mathbf{R}} \in \mathbb{R}^{B\times N}$ using the network to avoid the back-propagation of the null-space. 

\input{theroms/therom_1}

In addition to the primary task $F(\cdot)$, we utilize the self-supervised RGB reconstruction as the auxiliary task $G(\cdot)$ for test-time adaptation, as
\begin{equation}
\hat{\boldsymbol{\mathcal{I}}} = G (\boldsymbol{\mathcal{I}}, \hat{\mathbf{R}}),
\label{eqn:auxiliary}
\end{equation}
where the ground truth $\mathbf{R}$ is not needed. We empirically show in our experimental results that the auxiliary task also benefits the primary task, which coincides with the observation reported in~\cite{chi2021test}.

In this paper, $M = 1$ or $2$, and $\mathbf{I}_1$ and $\mathbf{I}_2$ represent a pair of RGB images from the same scene illuminated by a white LED $\mathbf{L}_1$ and an amber LED $\mathbf{L}_2$, respectively\footnote{Since the spectrum of an amber LED has a narrow band and is zero in most wavelengths, it can only serve as an auxiliary light source instead of the main one.}. The number of sampled spectral bands is 31 from 420nm to 720nm at 10nm increments. More detailed derivations are shown in the supplementary material.

\subsection{Architecture}
The overview of our proposed architecture is shown in Fig.~\ref{fig:arch}(a). It takes two RGB images $\mathbf{I}_1$ and $\mathbf{I}_2$ as inputs, and utilizes two separate conv layers to extract features. Feature maps from $\mathbf{I}_2$ are simply discarded for $M = 1$ or concatenated with those from $\mathbf{I}_1$ for $M = 2$. The channel size of the initial conv layers are set as 31 and are doubled/halved after downsampling/upsampling. All conv kernels are of size $3\times3$ and are followed by a LeakyReLU function~\cite{maas2013rectifier} except those before the concatenations, the element-wise operations ($+, -, \times$), and the outputs. The output channel size of the auxiliary task is 3 for $M = 1$ and 6 for $M = 2$.

We adopt an encoder-decoder architecture for SRR. Each scale of the encoder contains a conv layer followed by three resblocks. The decoder is similar but has an extra deconv layer to upscale the spatial dimension and a concatenation for skip-connection. We utilize four spectral-attention blocks~\cite{cai2022mst++} to extract spectral correlation after the encoder.

To explicitly estimate the CSS, we utilize the same encoder as the feature extractor and a conv layer to reduce the channel size to $3\times 31$, following which is a global average pooling layer. Despite using the same architecture, we do not share the parameters of the two encoders because we empirically observe that the network is difficult to converge. 

In the decoder, we adopt a pyramid scheme~\cite{dosovitskiy2015flownet, sun2018pwc, tao2018scale} by generating a spectral reflectance at the end of each scale, which can act as a “hint" for the prediction of finer scales. As shown in Fig.~\ref{fig:arch}(b), we downsample the RGB image stack $\boldsymbol{\mathcal{I}}$ with bilinear interpolation to match the spatial dimension at scale $i$ ($i\in\{1,2,3,4\}$) and calculate $\hat{\mathbf{R}}_{\hat{\boldsymbol{\mathcal{H}}}}^i$ with the estimated CSS $\hat{\mathbf{S}}$. The rescaling factor $\hat{\omega}^i$ is learned from the concatenation of $\hat{\mathbf{R}}_{\hat{\boldsymbol{\mathcal{H}}}}^i$ and $\Delta \hat{\mathbf{R}}^i$. $\hat{\mathbf{R}}^i$ is obtained by Eqn.~\ref{eqn:r_hat} and $\hat{\mathbf{R}}^1$ is our final recovered result $\hat{\mathbf{R}}$ in Eqn.~\ref{eqn:primary}.


A simple approach to fuse $\hat{\mathbf{R}}^i$ with features from scale $i-1$ is to directly upsample $\hat{\mathbf{R}}^i$ to scale $i-1$ with deconv layers and then concatenate them together~\cite{dosovitskiy2015flownet}. Nevertheless, the upsampled spectral reflectance lacks high-frequency information which needs further refinement. Inspired by the generalized Laplacian pyramid algorithm~\cite{aiazzi2002context} that fuses a high-resolution panchromatic image with a low-resolution multispectral image by feeding the weighted high-frequency information from the panchromatic image to the multispectral image, we propose a new \textbf{F}eature-g\textbf{U}ided up\textbf{S}ampling modul\textbf{E} (FUSE) that utilizes the feature $e^{i-1}$ from scale $i-1$ of the encoder to guide the upsampling. As shown in Fig.~\ref{fig:arch}(c), we exploit a downsampling-upsampling scheme to get the low-pass components of $e^{i-1}$ and then subtract it from $e^{i-1}$ for high-pass components $e^{i-1}_{high}$. The remaining low-pass components are concatenated with the upsampled recovery output $\hat{\mathbf{R}}^i$ and $e^{i-1}$ to extract local correlation, and generate local gain factor $m^i$ to reweight high-pass components which supplement the upsampled $\hat{\mathbf{R}}^i$ for refinement.

Most parameters of the two tasks are shared. As shown in Fig.~\ref{fig:arch}(a), we separate the parameters $\theta$ of the whole network into three components, $\theta_S$, $\theta_{Pri}$ and $\theta_{Aux}$, where $\theta_S$ represents the shared parameters, $\theta_{Pri}$ and $\theta_{Aux}$ represent the task-specific parameters for the primary task and the auxiliary task, respectively. We feed the output of the last shared resblock into two branches, one for generating the spectral reflectance $\hat{\mathbf{R}}$ (primary task), and the other with $\hat{\mathbf{R}}$ as an extra input to reconstruct the original RGB images as in Eqn.~\ref{eqn:auxiliary} (auxiliary task), so that the parameters of the primary task can be updated with only the auxiliary loss during test time. We adopt the L1 loss for both tasks as
\begin{equation}
\mathcal{L}_{Pri}(\theta_S, \theta_{Pri}) = \left\|\mathbf{S} - \hat{\mathbf{S}}\right\|_1 + \sum_{i=1}^{4}\left\|\mathbf{R}^i - \hat{\mathbf{R}}^i\right\|_1,
\label{eqn:p_loss}
\end{equation}
\begin{equation}
\mathcal{L}_{Aux}(\theta_S, \theta_{Pri}, \theta_{Aux}) = \left\|\boldsymbol{\mathcal{I}} - \hat{\boldsymbol{\mathcal{I}}}\right\|_1.
\label{eqn:a_loss}
\end{equation}

Directly updating the randomly initialized parameters with meta-auxiliary learning is time-consuming and unstable. Hence, we first initialize all the parameters by pre-training with the summation of the primary and the auxiliary losses following~\cite{chi2021test}, which is formulated as
\begin{equation}
\mathcal{L}_{Pre}(\theta) = \mathcal{L}_{Pri}(\theta_S, \theta_{Pri}) + \mathcal{L}_{Aux}(\theta_S, \theta_{Pri}, \theta_{Aux}).
\label{eqn:pre_loss}
\end{equation}

\subsection{Meta-auxiliary Learning}

\input{algorithms/algorithm_1}

The goal of meta-learning is to learn a general model for different tasks, which is able to rapidly adapt to new tasks with only a few steps~\cite{finn2017model}. In our case, we regard each triple $(\boldsymbol{\mathcal{I}}^k, \mathbf{S}^k, \mathbf{R}^k)$ ($k$ represents the index) as a task\footnote{To distinguish from the primary and auxiliary tasks, we utilize “meta-task" in the following text.} $\mathcal{T}^k$ of meta-learning. 

\mysubsubsection{Meta-auxiliary training} Given a meta-task $\mathcal{T}^k$, we first adapt the pre-trained parameters $\theta$ using several gradient descent updates based on only the auxiliary loss
\begin{equation}
\theta^k = \theta - \alpha\nabla_\theta\mathcal{L}_{Aux}^{\mathcal{T}^k}(\theta_S, \theta_{Pri}, \theta_{Aux}),
\label{eqn:adap_update}
\end{equation}
where $\alpha$ represents the adaptation learning rate. The update of Eqn.~\ref{eqn:adap_update} includes all the parameters with only $\boldsymbol{\mathcal{I}}^k$ utilized. 

The key of making the pre-trained parameters $\theta$ suitable for test-time adaptation is to update $\theta_S$ and $\theta_{Pri}$ of the primary task in the direction of minimizing the auxiliary loss. Thus, the meta-objective can be defined as 
\begin{equation}
\arg\min_{\theta_S, \theta_{Pri}}\sum_{k = 1}^{K} \mathcal{L}_{Pri}^{\mathcal{T}^k} (\theta_S^k, \theta_{Pri}^k),
\label{eqn:meta_obj}
\end{equation}
where $K$ is the number of sampled meta-tasks. The meta-optimization is then performed on Eqn.~\ref{eqn:meta_obj} via stochastic gradient descent
\begin{equation}
\theta \leftarrow \theta - \beta\sum_{k=1}^{K}\nabla_\theta\mathcal{L}_{Pri}^{\mathcal{T}^k}(\theta_S^k, \theta_{Pri}^k),
\label{eqn:meta_update}
\end{equation}
where $\beta$ represents the meta-learning rate. Note that the gradient in Eqn.~\ref{eqn:meta_update} is calculated based on $\theta^k$ but updates the original $\theta$ in Eqn.~\ref{eqn:adap_update}. The full algorithm is demonstrated in Alg.~\ref{alg:1}. Only $\theta_S$ and $\theta_{Pri}$ are updated in the outer loop, and $\theta_{Aux}$ is updated in the inner loop as
\begin{equation}
\theta_{Aux} \leftarrow \theta_{Aux} - \alpha\nabla_\theta\mathcal{L}_{Aux}^{\mathcal{T}^k}(\theta_{Aux}).
\label{eqn:aux_update}
\end{equation}

\input{algorithms/algorithm_2}

\mysubsubsection{Test-time adaptation} At test-time, we simply fine-tune the meta-learned parameters on a testing $\boldsymbol{\mathcal{I}}$ with Eqn.~\ref{eqn:adap_update} using several steps of gradient descent as shown in Alg.~\ref{alg:2}.

%% file: figures/spectrum.tex
\begin{figure}
\centering
\includegraphics[width=0.47\textwidth]{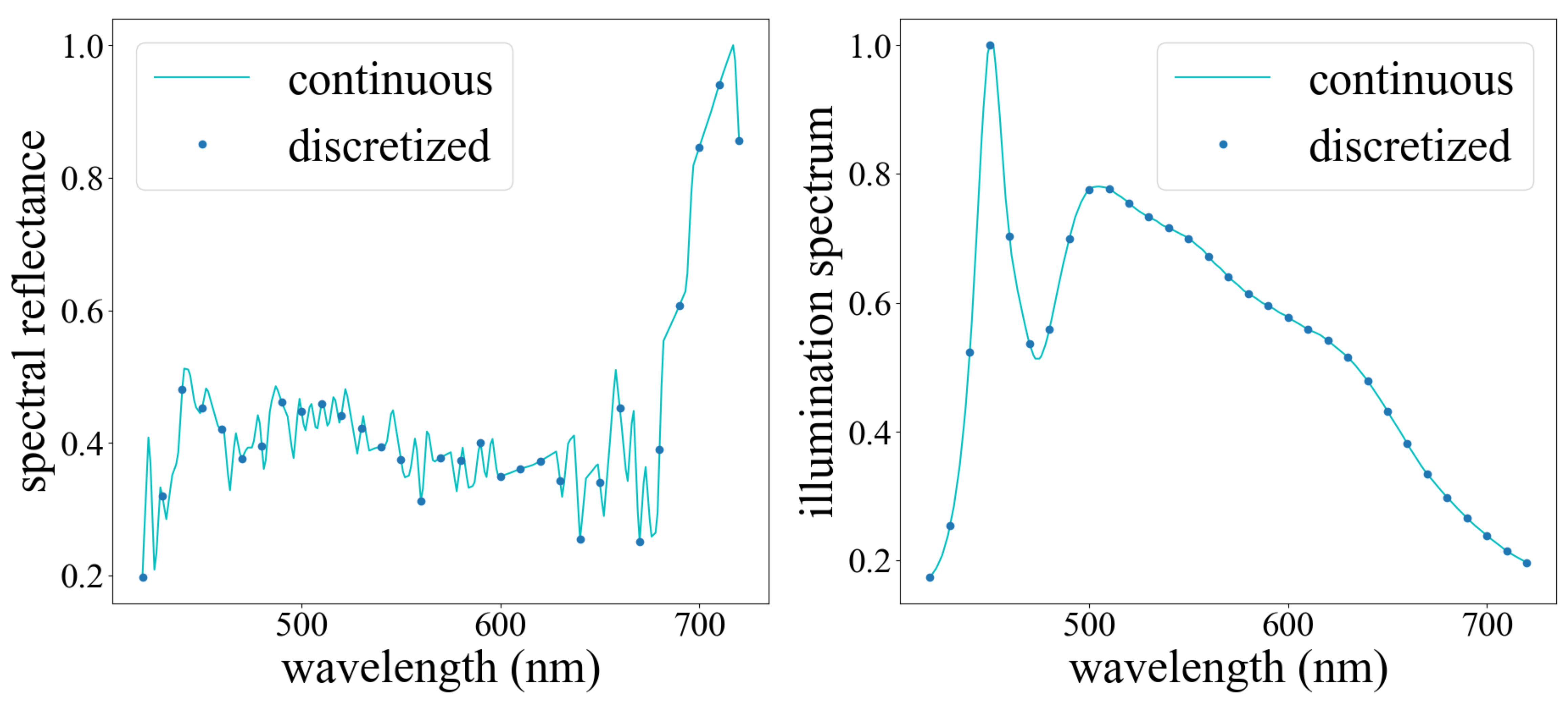}
\caption{The left and right figures show a spectral reflectance curve and the illumination spectrum of a white LED, respectively. We can see that discretization loses high-frequency information.}
\label{fig:spectrum}
\end{figure}

%% file: figures/architecture.tex
\begin{figure*}[t]
\centering
\includegraphics[width=\textwidth]{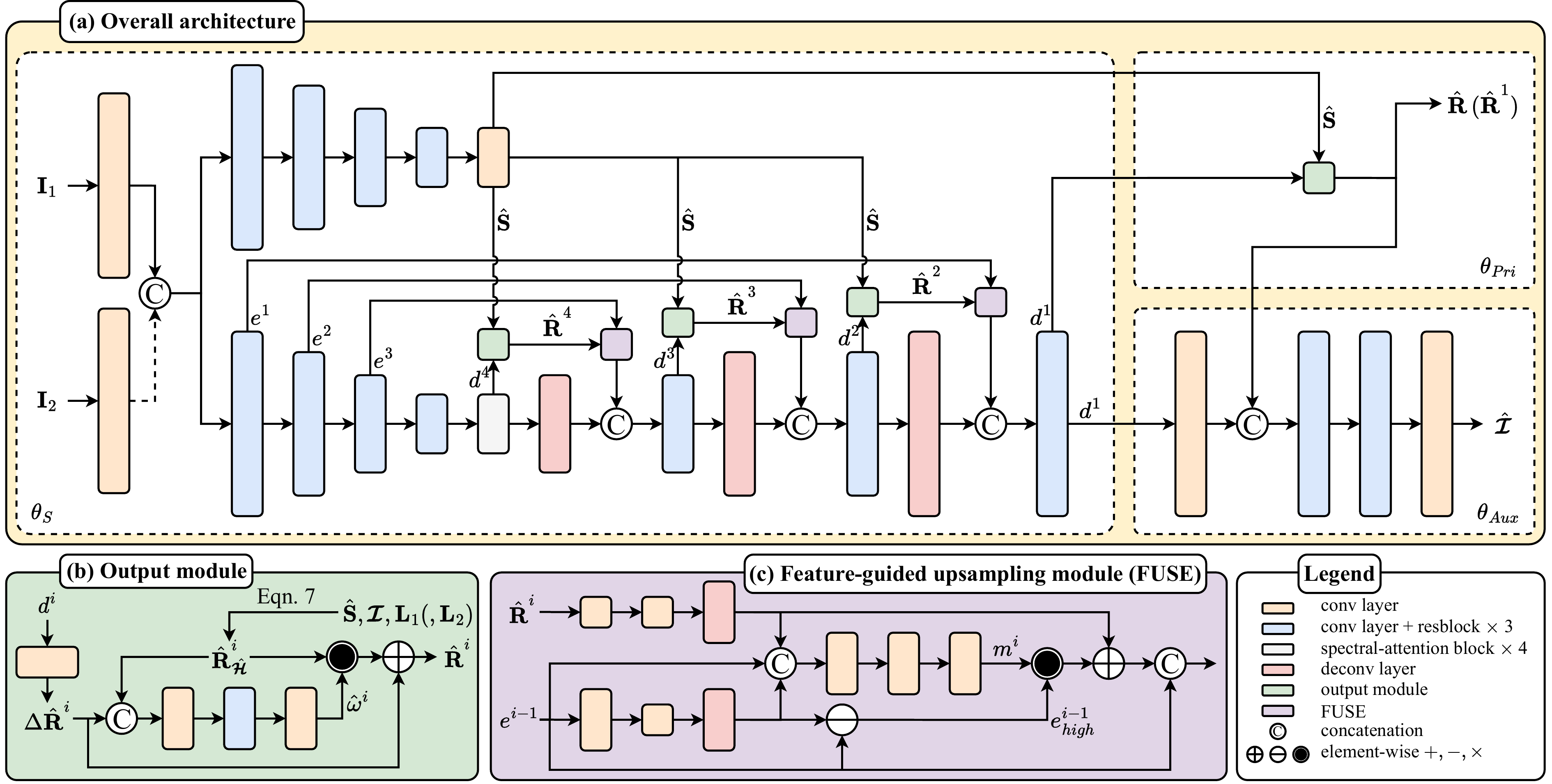}
\caption{Our proposed network architecture for SRR and meta-auxiliary learning. $e^i$ and $d^i$ denote the feature map from the encoder and the decoder, respectively, of scale $i$ ($i\in \{1, 2, 3, 4\}$), $\hat{\mathbf{R}}^i$ is the recovered reflectance of scale $i$ and $\hat{\mathbf{R}}^1$ represents the final recovered result $\hat{\mathbf{R}}$. The RGB image stack $\boldsymbol{\mathcal{I}}$ is downsampled to the corresponding scale before calculating $\hat{\mathbf{R}}_{\hat{\boldsymbol{\mathcal{H}}}}^i$. $\theta_{Pri}$ and $\theta_{Aux}$ denote the task-specific parameters for the primary task and the auxiliary task, respectively, and $\theta_S$ denotes the shared parameters. Our network consists of an encoder network to estimate the CSS, an encoder-decoder architecture for SRR, four spectral-attention layers to extract spectral correlation, output modules to generate $\hat{\mathbf{R}}^i$, and feature-guided upsampling modules (FUSEs) to upsample $\hat{\mathbf{R}}^i$ with the guidance of $e^{i - 1}$. The global average pooling before $\hat{\mathbf{S}}$ is omitted to simplify the illustration.}
\label{fig:arch}
\end{figure*}

%% file: theroms/therom_1.tex
\begin{thm} 
\label{the:theorem_1}
All possible solutions of $\hat{\mathbf{R}}$ share the same $\hat{\omega} \hat{\mathbf{R}}_{\hat{\boldsymbol{\mathcal{H}}}}$ component.
\end{thm}
\begin{proof} 
Let $\Delta\boldsymbol{\mathcal{I}}$ be the lost information of RGB images by discretization, $\Delta \boldsymbol{\mathcal{H}}$ be the difference between $\boldsymbol{\mathcal{H}}$ and $\hat{\boldsymbol{\mathcal{H}}}$, and $\Delta \omega$ be the difference between $\omega$ and $\hat{\omega}$. $\hat{\mathbf{R}}$ can be rewritten as
\begin{align}
\hat{\mathbf{R}}=&\hat{\mathbf{R}}^\parallel + \hat{\mathbf{R}}^\bot \nonumber\\
=& (\hat{\omega} + \Delta\omega) (\hat{\boldsymbol{\mathcal{H}}} + \Delta \boldsymbol{\mathcal{H}})^T \nonumber\\
&\cdot ((\hat{\boldsymbol{\mathcal{H}}} + \Delta \boldsymbol{\mathcal{H}}) \cdot (\hat{\boldsymbol{\mathcal{H}}}+ \Delta \boldsymbol{\mathcal{H}})^T)^{-1}
\cdot (\boldsymbol{\mathcal{I}} - \Delta\boldsymbol{\mathcal{I}}) + \hat{\mathbf{R}}^\bot \nonumber\\
=& \hat{\omega} \underbrace{\hat{\boldsymbol{\mathcal{H}}}^T \cdot (\hat{\boldsymbol{\mathcal{H}}} \cdot \hat{\boldsymbol{\mathcal{H}}}^T)^{-1} \cdot \boldsymbol{\mathcal{I}}}_{\hat{\mathbf{R}}_{\hat{\boldsymbol{\mathcal{H}}}}} + \Delta \hat{\mathbf{R}}.
\label{eqn:r_hat}
\end{align}
\hfill $\blacksquare$
\end{proof}

%% file: algorithms/algorithm_1.tex
\begin{algorithm}[t]
	\DontPrintSemicolon
	\SetAlgoLined
	
	\KwInput{$(\boldsymbol{\mathcal{I}}, \mathbf{S}, \mathbf{R})$ triples\\
	\,\,\,\,\,\,\,\,\,\,\,\,\,\,\,\,\,\,$\alpha$, $\beta$: learning rates}
	\KwOutput{$\theta$: meta-auxiliary trained parameters}
	Randomly initialize $\theta$, $\theta=\{\theta_S, \theta_{Pri}, \theta_{Aux}\}$\\
	\While{not converged}
	{
		Sample a batch of triples $\{\boldsymbol{\mathcal{I}}^k, \mathbf{S}^k, \mathbf{R}^k\}_{k=1}^K$\\
		Evaluate pre-training loss $\mathcal{L}_{Pre}$ by Eqn.~\ref{eqn:pre_loss}\\
		Update $\theta$ with respect to $\mathcal{L}_{Pre}$
		
	}

	\While{not converged}
	{
		Sample a batch of triples $\{\boldsymbol{\mathcal{I}}^k, \mathbf{S}^k, \mathbf{R}^k\}_{k=1}^K$\\
		\For{each $k$}{
			Evaluate auxiliary loss $\mathcal{L}_{Aux}$ by Eqn.~\ref{eqn:a_loss} \\
			Compute adapted parameters $\theta^k$ with gradient descent by Eqn.~\ref{eqn:adap_update}\\
			Update $\theta_{Aux}$ by Eqn.~\ref{eqn:aux_update}
		}
		Update $\theta_S$ and $\theta_{Pri}$ by Eqn.~\ref{eqn:meta_update}
	}
	\caption{Meta-auxiliary Training}
	\label{alg:1}
\end{algorithm}

%% file: algorithms/algorithm_2.tex
\begin{algorithm}[t]
	\DontPrintSemicolon
	\SetAlgoLined
	
	\KwInput{A testing RGB image stack $\boldsymbol{\mathcal{I}}$\\
	\,\,\,\,\,\,\,\,\,\,\,\,\,\,\,\,\,\,$n$: number of gradient updates\\
	\,\,\,\,\,\,\,\,\,\,\,\,\,\,\,\,\,\,$\alpha$: adaptation learning rate}
	\KwOutput{Recovered spectral reflectance $\hat{\mathbf{R}}$}
	Initialize network parameters with meta-learned $\theta$\\
	\For{$n$ steps}
	{
		Evaluate auxiliary loss $\mathcal{L}_{Aux}$ by Eqn.~\ref{eqn:a_loss} \\
		Update $\theta \leftarrow \theta - \alpha\nabla_\theta\mathcal{L}_{Aux}(\theta_S, \theta_{Pri}, \theta_{Aux})$
		
	}
	\Return $\hat{\mathbf{R}}$ from Eqn.~\ref{eqn:primary}
	
	\caption{Test-time Adaptation}
	\label{alg:2}
\end{algorithm}

%% file: 4_experiments.tex
\input{tables/quantitative}

\section{Experiments}
\subsection{Data Preparation and Evaluation Metrics}
\mysubsubsection{Synthetic data} TokyoTech~\cite{monno2018single} contains 16 spectral reflectance images from 420nm to 1000nm at 10nm increments, and we utilize the first 31 bands. ICVL~\cite{arad2016sparse} contains 201 hyperspectral images under daylight illumination from 400nm to
1000nm at 1.5nm increments. We divide the hyperspectral images by the daylight illumination spectrum~\cite{lspdd} to simulate the spectral reflectance, then downsample from 420nm to 720nm at 10nm increments. We randomly select $75\%$ images from two datasets for training and the rest for testing. Jiang~\etal~\cite{jiang2013space} provide 28 CSSs and we randomly select 23 for generating training inputs and the rest for testing. The illumination spectra of white and amber LEDs are collected with a Specim IQ mobile hyperspectral camera and are downsampled using the same scheme. We normalize two illumination spectra to the range $[0, 1]$ and keep their relative intensity. To simulate the continuous spectra, we interpolate the spectral reflectance spectra, CSSs and illumination spectra at 1nm increments before generating RGB images with Eqn.~\ref{eqn:mm}. 

\mysubsubsection{Real data} To evaluate the robustness of models trained on synthetic data, we collect 25 spectral reflectance images with a Specim IQ and the corresponding RGB images under white and amber LEDs with a Canon 6D camera which is not included in the training data. The illumination spectra are represented as the spectral radiance of a white reference panel under two LEDs. The reflectance spectra are downsampled from 420nm to 720nm at 10nm increments. We first convert the downsampled spectra to RGB using a randomly selected CSS from~\cite{jiang2013space}, then we adopt feature matching with SIFT features~\cite{lowe1999object} to align the images of two cameras. Note that images without enough features for matching are removed. Feasibility analysis of data capture in real world is shown in the supplementary materials.

\mysubsubsection{Evaluation metrics} We adopt the mean absolute error (MAE), rooted mean square error (RMSE), spectral angle similarity (SAS~\cite{dennison2004comparison}), peak signal-to-noise ratio (PSNR~\cite{korhonen2012peak}) and structural similarity (SSIM~\cite{wang2004image}) as the metrics to evaluate the performance of SRR.

\subsection{Implementation Details} All images are linearly rescaled to the range $[0, 1]$. Training images are cropped into 128$\times$128 patches with a stride of 64, and are augmented by random flips. The batch size is set to 64. We adopt the Adam optimizer~\cite{kingma2014adam} for pre-training with a learning rate $10^{-4}$ and the Cosine Annealing scheme~\cite{loshchilov2016sgdr} for 300 epochs. During the meta-auxiliary learning, we set $\alpha$ and $\beta$ to $1\times10^{-2}$ and $5\times10^{-5}$, respectively. For test-time adaptation, we perform $n=5$ gradient descent updates. All experiments are conducted on a single NVIDIA RTX A6000 GPU with 48GB of RAM.

\subsection{Quantitative Evaluations}
\label{sec:quan}
We first evaluate the performance of $M = 1$. We compare our method with 6 state-of-the-art methods for spectral reconstruction from a single RGB image, including HSCNN+~\cite{shi2018hscnn+}, MSDCNN~\cite{yan2018accurate}, PADFMN~\cite{zhang2020pixel}, QDO~\cite{li2022quantization}, MST++~\cite{cai2022mst++}, and DRCRN~\cite{li2022drcr}. For fair comparison, we remove the DOE optimization of QDO. All of these competing methods are retrained with our selected synthetic data. The evaluation results are listed in the first part of Tab.~\ref{tab:quan}. We can see that our proposed architecture outperforms other methods even with only the pre-trained model, and the MAXL obviously improves the performance especially on the challenging real data (0.65dB), which demonstrates the importance of utilizing internal information. Since our model is trained on the synthetic data, it is reasonable that the performance gain of MAXL on the synthetic data is not as much as that on the real data. 




We also evaluate the effectiveness of the extra illumination ($M = 2$). As reported in the second part of Tab.~\ref{tab:quan}, it demonstrates 0.63dB and 2.39dB improvement over  $M = 1$ on synthetic data and real data, respectively. 

The evaluation of computational complexity on images of size $1392\times 1303$ is shown in Tab.~\ref{tab:comp}. We can see that our method without MAXL is faster than most of the other methods with comparable number of parameters, and the test-time adaptation only takes seconds.

\input{tables/complexity}

\input{tables/complete_ablation_1}
\input{tables/complete_ablation_2}

\input{tables/ablation_3}
\input{tables/ablation_4}

\subsection{Qualitative Evaluations}
\label{sec:qual}
The qualitative comparison results of the 630nm band of the spectral reflectance are shown in Fig.~\ref{fig:qualitative}. The RGB images under white LED, the ground truth, and the error maps of all competing methods are shown from top to bottom. The first four columns and last three columns show the results from synthetic data and real data, respectively. We can see that our method with MAXL performs better than others and is more robust on real data. More qualitative evaluations are shown in the supplementary materials.

Fig.~\ref{fig:real_maxl} visualizes the effect of using one ($M=1$) or two ($M=2$) illuminations  with/without MAXL on real data. It shows that the extra illumination can help to reduce the overall error of the entire image, and the MAXL benefits some local details.

To evaluate the generated CSSs of five selected testing cameras, we display the visual comparison between the ground truth and our estimation in Fig.~\ref{fig:crf}. It demonstrates that our proposed method can accurately estimate the CSSs that are unseen during the training.

\input{figures/qualitative}
\input{figures/real_maxl}
\input{figures/crf}

\subsection{Ablation Studies}\label{sec:ablation}

We conduct ablation studies on the synthetic data. As shown in Tab.~\ref{tab:com_ablation_1}, pyramid learning (multi-scale outputs) plays a vital role in the performance, and a proper fusion strategy is also important compared with no FUSE (simply output encoder feature $e^{i - 1}$ in FUSE) and zero $m^i$. We also remove $\hat{\mathbf{R}}_{\hat{\boldsymbol{\mathcal{H}}}}^i$ (use $\Delta\hat{\mathbf{R}}^i$ as $\hat{\mathbf{R}}^i$), $\Delta\hat{\mathbf{R}}^i$ (use $\hat{\omega}^i\hat{\mathbf{R}}_{\hat{\boldsymbol{\mathcal{H}}}}^i$ as $\hat{\mathbf{R}}^i$) and $\hat{\omega}^i$ (use $\hat{\mathbf{R}}_{\hat{\boldsymbol{\mathcal{H}}}}^i + \Delta\hat{\mathbf{R}}^i$ as $\hat{\mathbf{R}}^i$) in all output modules to investigate the impact of the subspace component. We can see that the physical properties of the spectral reflectance can benefit its recovery, but its subspace component alone is insufficient. Utilizing the ground truth CSSs to calculate the $\hat{\boldsymbol{\mathcal{H}}}$ can further improve the results. Besides, the performance gain from the spectral-attention blocks illustrates the effectiveness of spectral correlation. The experiments without the auxiliary task also demonstrate that it can help the optimization of the primary task. 

\input{figures/application}

As shown in Tab.~\ref{tab:com_ablation_2}, after fine-tuning the pre-trained model with meta-auxiliary training, the evaluation results show an improvement but are still sub-optimal. We also evaluate the performance of direct test-time adaptation without meta-auxiliary training. While the performance improvement is minor, we do not observe the catastrophic forgetting as mentioned in~\cite{chi2021test}. 

We also investigate the effect of gradient descent update step $n$ as reported in Tab.~\ref{tab:ablation_3}. We choose $n = 5$ for the best performance. More update steps may lead to the overfitting on the auxiliary task. Note that we utilize the same $n$ during training and testing.

In Section~\ref{sec:quan}, we illustrate the performance of using one ($M=1$) or two ($M=2$) illuminations. To further demonstrate the robustness of our proposed architecture with more illuminations, we utilize the spectrum of a halogen light as the illumination $\mathbf{L}_3$ to synthesize the input RGB image $\mathbf{I}_3$. Reported in Tab.~\ref{tab:ablation_4} are the results with $1\sim 3$ illuminations. As we can see, utilizing a third illumination can further improve the performance of recovery.

\input{figures/limitation}

\subsection{Applications}

As mentioned in Section~\ref{sec:intro}, spectral reflectance describes the distinctive intrinsic characteristics of an object’s material or composition and is widely leveraged for material recognition~\cite{li2019deep, bioucas2013hyperspectral, thenkabail2013selection}. For example, it has been found to be a more reliable cue for assessing the quality of food, particularly fruits, compared to RGB images~\cite{wang2016recent}. To demonstrate that our recovered spectral reflectance possesses the same property, we conduct experiments of distinguishing between salt and sugar, and detecting fruit puncture in a tomato. The objects in each case have similar RGB colors, with salt and sugar both appearing white, and the tomato peel and pulp both appearing red. 

Fig.~\ref{fig:application} shows the two example results. We can see that the discrepancy of different materials (salt and sugar, tomato peel and pulp) are more distinguishable on our recovered spectral reflectance than on the original RGB image. For example, the error between the salt and sugar on RGB images is only $1.78\times 10^{-5}$ but $0.53$ on the recovered spectral reflectance, and the error between the tomato peel and pulp on RGB images and spectral reflectance are $0.09$ and $0.17$, respectively.

\subsection{Limitations}
Our method is limited on bands that have little impact on the RGB images, such as marginal bands, which is a common issue for most approaches. As shown in Fig.~\ref{fig:spectrum} and Fig.~\ref{fig:crf}, the illumination spectra and CSSs are heterogeneous, and the intensity of marginal bands (\eg, 430nm) is much lower than that of central bands (\eg, 600nm). As a result, marginal bands are harder to recover. The error maps of our recovered 430nm and 600nm bands are shown in Fig.~\ref{fig:limitation}, which illustrate that the errors on the marginal bands are higher than that of central bands.

%% file: tables/quantitative.tex
\begin{table*}[t]
\centering
\renewcommand\arraystretch{1.5}
\setlength\tabcolsep{3pt}
\caption{Quantitative evaluations. All compared methods are trained on the synthetic data. Ours and Ours$\dagger$ represent the $M = 1$ (white LED only) and $M = 2$ (white$\&$amber LEDs), respectively. “pre-trained" represents the model without meta-auxiliary training and test-time adaptation. } 
\begin{tabular}{lcccccccccc}
\toprule
\multirow{2}{*}{Methods} & \multicolumn{5}{c}{Synthetic data}                                                                                       & \multicolumn{5}{c}{Real data}                                                                                            \\ 
\cmidrule[0.25pt](lr){2-6} \cmidrule[0.25pt](lr){7-11} 
                         & \multicolumn{1}{c}{MAE$\downarrow$}   & \multicolumn{1}{c}{RMSE$\downarrow$}  & \multicolumn{1}{c}{SAS$\downarrow$}   & \multicolumn{1}{c}{PSNR$\uparrow$}  & SSIM$\uparrow$  & \multicolumn{1}{c}{MAE$\downarrow$}   & \multicolumn{1}{c}{RMSE$\downarrow$}  & \multicolumn{1}{c}{SAS$\downarrow$}   & \multicolumn{1}{c}{PSNR$\uparrow$}  & SSIM$\uparrow$  \\ 
\midrule
HSCNN+~\cite{shi2018hscnn+}                   & \multicolumn{1}{c}{0.1261} & \multicolumn{1}{c}{0.1594} & \multicolumn{1}{c}{0.1418} & \multicolumn{1}{c}{16.96} & 0.7837 & \multicolumn{1}{c}{0.3107} & \multicolumn{1}{c}{0.3526} & \multicolumn{1}{c}{0.5521} & \multicolumn{1}{c}{9.11} & 0.3877\\ 
MSDCNN~\cite{yan2018accurate}                   & \multicolumn{1}{c}{0.0877} & \multicolumn{1}{c}{0.1124} & \multicolumn{1}{c}{0.1027} & \multicolumn{1}{c}{19.76} & 0.8400 & \multicolumn{1}{c}{0.3136} & \multicolumn{1}{c}{0.3563} & \multicolumn{1}{c}{0.5585} & \multicolumn{1}{c}{9.02} & 0.3821\\ 
PADFMN~\cite{zhang2020pixel}                   & \multicolumn{1}{c}{0.0851} & \multicolumn{1}{c}{0.1102} & \multicolumn{1}{c}{0.1010} & \multicolumn{1}{c}{20.15} & 0.8257 & \multicolumn{1}{c}{0.2746} & \multicolumn{1}{c}{0.3214} & \multicolumn{1}{c}{0.5217} & \multicolumn{1}{c}{9.93} & 0.3770 \\ 
QDO~\cite{li2022quantization}                    & \multicolumn{1}{c}{0.1494} & \multicolumn{1}{c}{0.1889} & \multicolumn{1}{c}{0.1295} & \multicolumn{1}{c}{15.14} & 0.7759 & \multicolumn{1}{c}{0.4665} & \multicolumn{1}{c}{0.5330} & \multicolumn{1}{c}{0.6139} & \multicolumn{1}{c}{5.52} & 0.2883\\ 
MST++~\cite{cai2022mst++}                    & \multicolumn{1}{c}{0.0724} & \multicolumn{1}{c}{0.0927} & \multicolumn{1}{c}{0.0865} & \multicolumn{1}{c}{21.72} & 0.8611 & \multicolumn{1}{c}{0.2400} & \multicolumn{1}{c}{0.2944} & \multicolumn{1}{c}{0.5312} & \multicolumn{1}{c}{10.69} & 0.3383\\ 
DRCRN~\cite{li2022drcr}                    & \multicolumn{1}{c}{0.0750} & \multicolumn{1}{c}{0.0998} & \multicolumn{1}{c}{0.0894} & \multicolumn{1}{c}{20.98} & 0.8429 & \multicolumn{1}{c}{0.2717} & \multicolumn{1}{c}{0.3154} & \multicolumn{1}{c}{0.5501} & \multicolumn{1}{c}{10.09} & 0.3992\\ 
Ours (pre-trained)       & \multicolumn{1}{c}{0.0625} & \multicolumn{1}{c}{0.0828} & \multicolumn{1}{c}{0.0748} & \multicolumn{1}{c}{22.91} & 0.8818 & \multicolumn{1}{c}{0.2313} & \multicolumn{1}{c}{0.2783} & \multicolumn{1}{c}{0.5174} & \multicolumn{1}{c}{11.19} & 0.4721\\ 
Ours                     & \multicolumn{1}{c}{\textbf{0.0607}} & \multicolumn{1}{c}{\textbf{0.0809}} & \multicolumn{1}{c}{\textbf{0.0734}} & \multicolumn{1}{c}{\textbf{23.09}} & \textbf{0.8833} & \multicolumn{1}{c}{\textbf{0.2136}} & \multicolumn{1}{c}{\textbf{0.2590}} & \multicolumn{1}{c}{\textbf{0.4934}} & \multicolumn{1}{c}{\textbf{11.84}} & \textbf{0.4947}\\ 
\midrule
Ours$\dagger$ (pre-trained)      & \multicolumn{1}{c}{0.0580} & \multicolumn{1}{c}{0.0778} & \multicolumn{1}{c}{0.0696} & \multicolumn{1}{c}{23.67} & 0.8891 & \multicolumn{1}{c}{0.1657} & \multicolumn{1}{c}{0.2137} & \multicolumn{1}{c}{0.4426} & \multicolumn{1}{c}{13.56} & 0.5641\\ 
Ours$\dagger$                    & \multicolumn{1}{c}{\textbf{0.0575}} & \multicolumn{1}{c}{\textbf{0.0771}} & \multicolumn{1}{c}{\textbf{0.0691}} & \multicolumn{1}{c}{\textbf{23.72}} & \textbf{0.8905} & \multicolumn{1}{c}{\textbf{0.1536}} & \multicolumn{1}{c}{\textbf{0.1997}} & \multicolumn{1}{c}{\textbf{0.4095}} & \multicolumn{1}{c}{\textbf{14.23}} & \textbf{0.5796}\\ 
\bottomrule
\end{tabular}
\label{tab:quan}
\end{table*}

%% file: tables/complexity.tex
\begin{table}[t]
\centering
\renewcommand\arraystretch{1.5}
\setlength\tabcolsep{3pt}
\caption{Evaluations of computational complexity. Ours and Ours$\dagger$ represent the $M = 1$ (white LED only) and $M = 2$ (white$\&$amber LEDs), respectively. “pre-trained" represents the model without meta-auxiliary training and test-time adaptation. All evaluations are calculated on images of size 1392$\times$1303.}
\begin{tabular}{lccc}
\toprule
Methods                       & \#Params   & FLOPs  &   Inference time    \\ 
\midrule
HSCNN+~\cite{shi2018hscnn+}             & \textbf{7.98}\boldmath{$\times10^{5}$} & 2.88$\times10^{12}$ & \textbf{0.020 sec} \\ 
MSDCNN~\cite{yan2018accurate}    & 2.67$\times10^{7}$     & 2.27$\times10^{12}$   & 0.023 sec  \\ 
PADFMN~\cite{zhang2020pixel}    & 3.17$\times10^{7}$  & 9.02$\times10^{12}$ & 0.334 sec  \\ 
QDO~\cite{li2022quantization}   & 1.47$\times10^{9}$  & 1.38$\times10^{12}$  & 0.308 sec \\ 
MST++~\cite{cai2022mst++}   & 1.62$\times10^{6}$  & \textbf{1.20}\boldmath{$\times10^{12}$} & 0.239 sec\\ 
DRCRN~\cite{li2022drcr}   & 9.48$\times10^{6}$  & 3.23$\times10^{13}$ & 0.538 sec \\ 
Ours (pre-trained)   & 2.41$\times10^{7}$  & 5.03$\times10^{12}$ & 0.145 sec  \\ 
Ours  & 2.41$\times10^{7}$  & 2.57$\times10^{13}$ & 6.018 sec \\ 
\midrule
Ours$\dagger$ (pre-trained)   & 2.42$\times10^{7}$  & 5.10$\times10^{12}$ & 0.153 sec \\ 
Ours$\dagger$   & 2.42$\times10^{7}$  & 2.61$\times10^{13}$ & 6.082 sec \\ 
\bottomrule
\end{tabular}
\vspace{-1em}
\label{tab:comp}
\end{table}

%% file: tables/complete_ablation_1.tex
\begin{table}[t]
\centering
\renewcommand\arraystretch{1.5}
\setlength\tabcolsep{3pt}
\caption{Ablation studies of network components.} 
\begin{tabular}{lccccc}
\toprule
Methods                       & MAE$\downarrow$   & RMSE$\downarrow$  & SAS$\downarrow$  & PSNR$\uparrow$  &  SSIM$\uparrow$   \\ 
\midrule
Ours (pre-trained)       & \textbf{0.0625} & \textbf{0.0828} & \textbf{0.0748} & \textbf{22.91} & \textbf{0.8818} \\ 
w/o pyramid       & 0.1277 & 0.1585 & 0.1432 & 16.58 & 0.7420 \\
w/o FUSE        & 0.0676 & 0.0887 & 0.0803 & 22.27 & 0.8738 \\
w/ zero $m^i$ in FUSE       & 0.0711 & 0.0922 & 0.0830 & 22.08 & 0.8696 \\
w/o $\hat{\mathbf{R}}_{\hat{\boldsymbol{\mathcal{H}}}}^i$       & 0.0669 & 0.0876 & 0.0798 & 22.55 & 0.8770 \\ 
w/o $\hat{\omega}^i$       & 0.0691 & 0.0909 & 0.0824 & 22.24 & 0.8726 \\
w/o $\Delta\hat{\mathbf{R}}^i$       & 0.3485 & 17.4572 & 0.8131 & 1.19 & 0.3836\\ 
w/ ground truth CSSs       & \textbf{0.0621} & \textbf{0.0824} & \textbf{0.0744} & \textbf{22.95} & \textbf{0.8818} \\ 
w/o spectral-attention       & 0.0730 & 0.0958 & 0.0894 & 21.44 & 0.8678 \\ 
w/o auxiliary task       & 0.0674 & 0.0888 & 0.0796 & 22.46 & 0.8703 \\ 
\bottomrule
\end{tabular}
\label{tab:com_ablation_1}
\end{table}

%% file: tables/complete_ablation_2.tex
\begin{table}[t]
\centering
\renewcommand\arraystretch{1.5}
\setlength\tabcolsep{3pt}
\caption{Ablation studies of learning strategies.} 
\begin{tabular}{lccccc}
\toprule
Methods                       & MAE$\downarrow$   & RMSE$\downarrow$  & SAS$\downarrow$  & PSNR$\uparrow$  &  SSIM$\uparrow$   \\ 
\midrule
Ours (pre-trained)       & 0.0625 & 0.0828 & 0.0748 & 22.91 & 0.8818 \\ 
w/ meta-auxiliary training        & 0.0611 & 0.0814 & 0.0739 & 23.03 & 0.8828 \\
w/ test-time adaptation        & 0.0624 & 0.0827 & 0.0745 & 22.92 & 0.8822 \\ 
w/ MAXL                    & \textbf{0.0607} & \textbf{0.0809} & \textbf{0.0734} & \textbf{23.09} & \textbf{0.8833}\\ 
\bottomrule
\end{tabular}
\label{tab:com_ablation_2}
\end{table}

%% file: tables/ablation_3.tex
\begin{table}[t]
\centering
\renewcommand\arraystretch{1.5}
\setlength\tabcolsep{3pt}
\caption{Ablation studies of number of gradient descent updates $n$. }
\begin{tabular}{lccccc}
\toprule
Methods                       & MAE$\downarrow$   & RMSE$\downarrow$  & SAS$\downarrow$  & PSNR$\uparrow$  &  SSIM$\uparrow$   \\ 
\midrule
$n = 0$                   & 0.0625 & 0.0828 & 0.0748 & 22.91 & 0.8818 \\
$n = 1$                   & 0.0613 & 0.0816 & 0.0737 & 23.01 & 0.8826\\ 
$n = 2$                  & 0.0612 & 0.0812 & 0.0736 & 23.02 & 0.8828 \\ 
$n = 3$                  & 0.0611 & 0.0814 & 0.0736 & 23.03 & 0.8828 \\ 
$n = 4$                   & 0.0611 & 0.0812 & 0.0735 & 23.03 & 0.8830 \\ 
$n = 5$               & \textbf{0.0607} & \textbf{0.0809} & \textbf{0.0734} & \textbf{23.09} & \textbf{0.8833} \\
$n = 6$                  & 0.0609 & 0.0813 & \textbf{0.0734} & 23.07 & 0.8827 \\ 
\bottomrule
\end{tabular}
\label{tab:ablation_3}
\end{table}

%% file: tables/ablation_4.tex
\begin{table}[t]
\centering
\renewcommand\arraystretch{1.5}
\setlength\tabcolsep{3pt}
\caption{Ablation studies of number of different illuminations $M$. }
\begin{tabular}{lccccc}
\toprule
Methods                       & MAE$\downarrow$   & RMSE$\downarrow$  & SAS$\downarrow$  & PSNR$\uparrow$  &  SSIM$\uparrow$   \\ 
\midrule
$M = 1$                   & 0.0625 & 0.0828 & 0.0748 & 22.91 & 0.8818 \\
$M = 2$                   & 0.0580 & 0.0778 & 0.0696 & 23.67 & 0.8891 \\
$M = 3$                   & \textbf{0.0555} & \textbf{0.0742} & \textbf{0.0674} & \textbf{23.97} & \textbf{0.8939} \\
\bottomrule
\end{tabular}
\label{tab:ablation_4}
\end{table}

%% file: figures/qualitative.tex
\begin{figure*}[!htbp]
\centering
\includegraphics[width=0.98\textwidth]{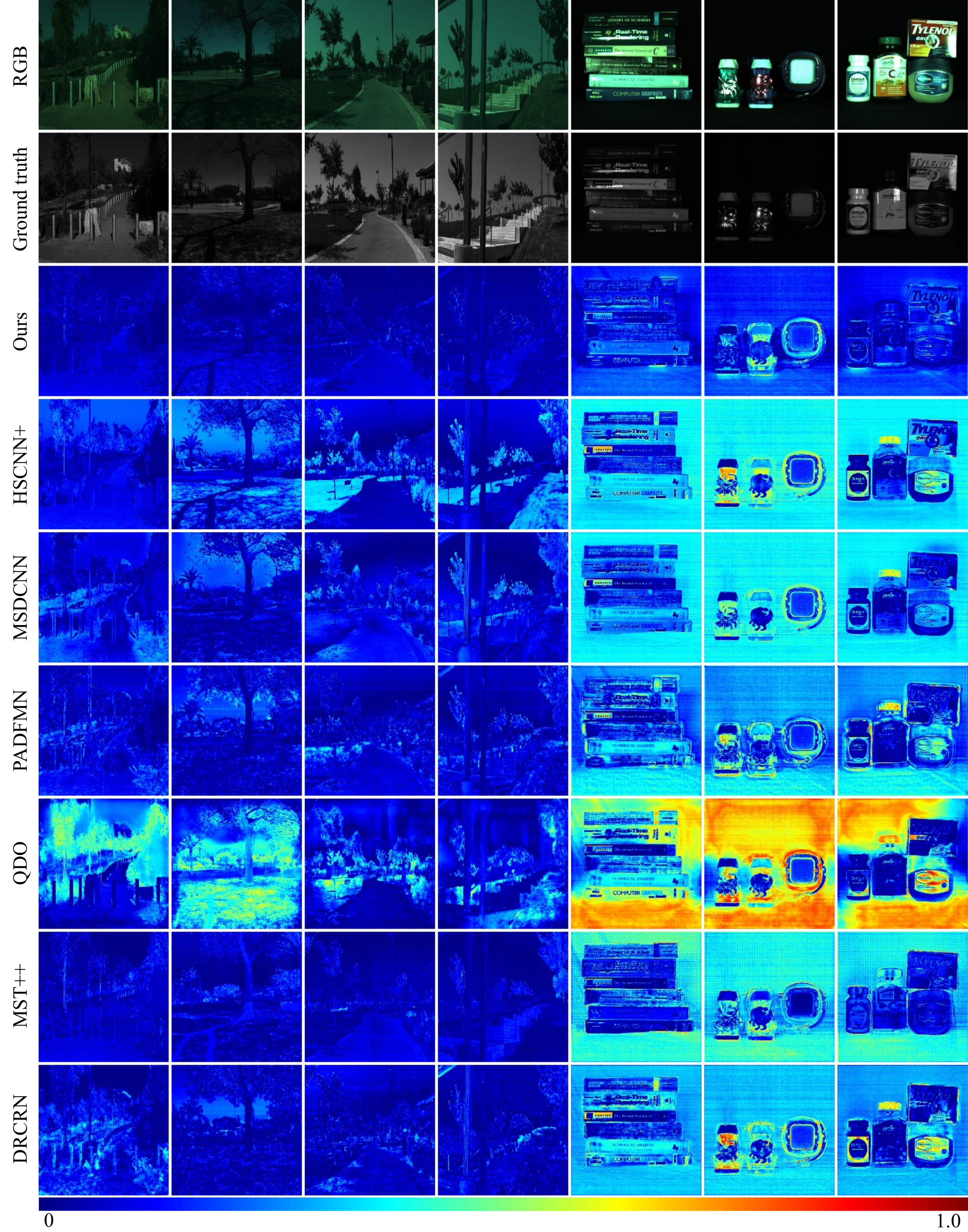}
\caption{Qualitative comparison of error maps (MAE between the recovered results and the ground truth) with state-of-the-art approaches. The first four columns are from the synthetic data and last three columns are from our collected real data. }
\label{fig:qualitative}
\end{figure*}

%% file: figures/real_maxl.tex
\begin{figure*}[!htbp]
\centering
\includegraphics[width=\textwidth]{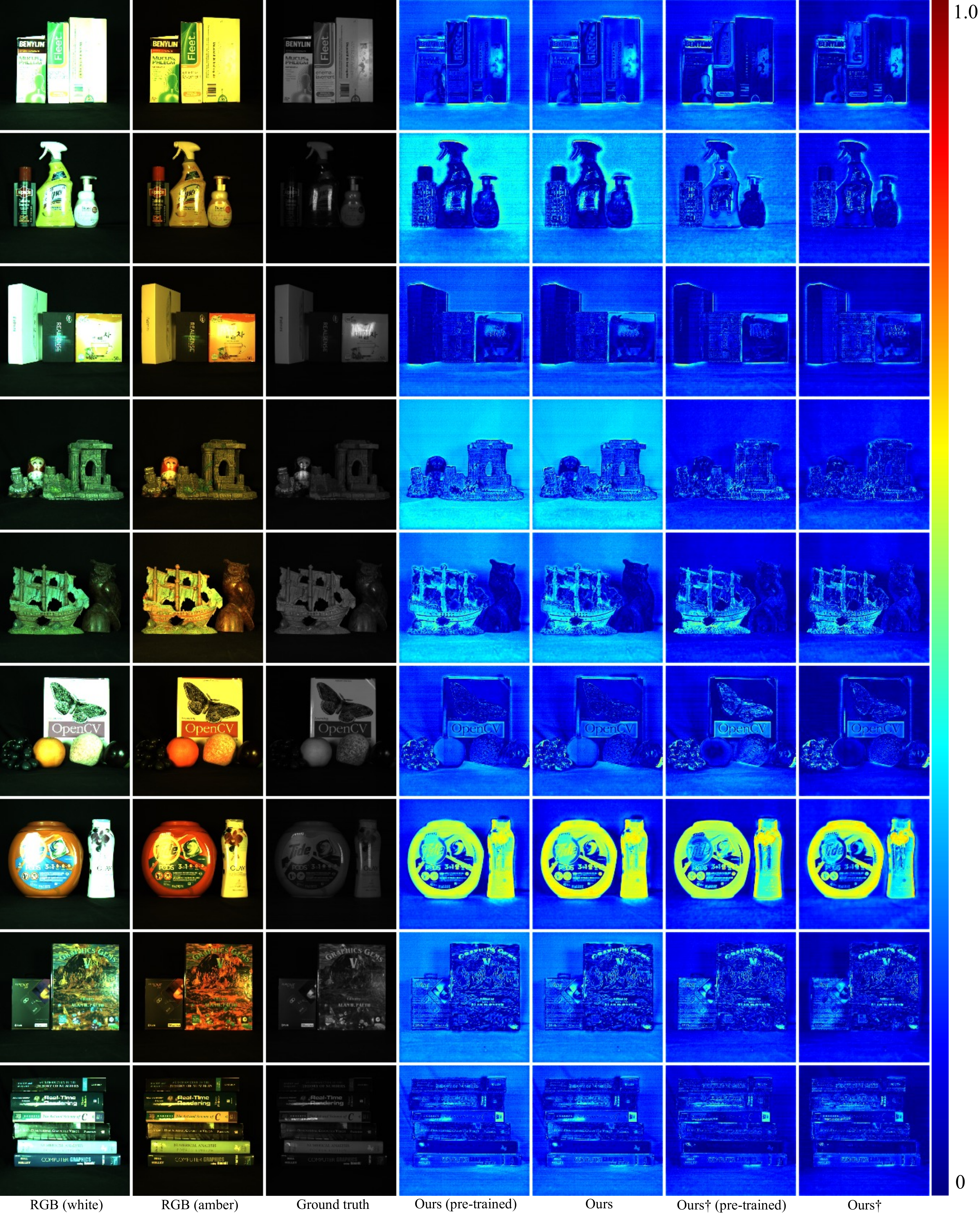}
\caption{Qualitative comparison of error maps (MAE between the recovered results and the ground truth) of our method with/without MAXL for $M = 1$ and $M = 2$ on real data. }
\label{fig:real_maxl}
\end{figure*}

%% file: figures/crf.tex
\begin{figure*}[t]
\centering
\includegraphics[width=0.95\textwidth]{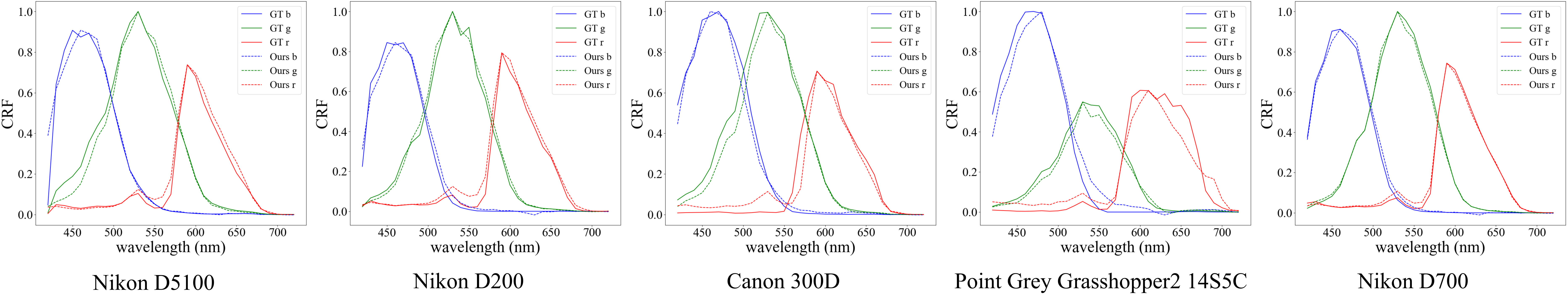}
\caption{Visual comparison of the ground truth and our estimated CSSs.}
\label{fig:crf}
\end{figure*}

%% file: figures/application.tex
\begin{figure}[t]
\centering
\includegraphics[width=0.47\textwidth]{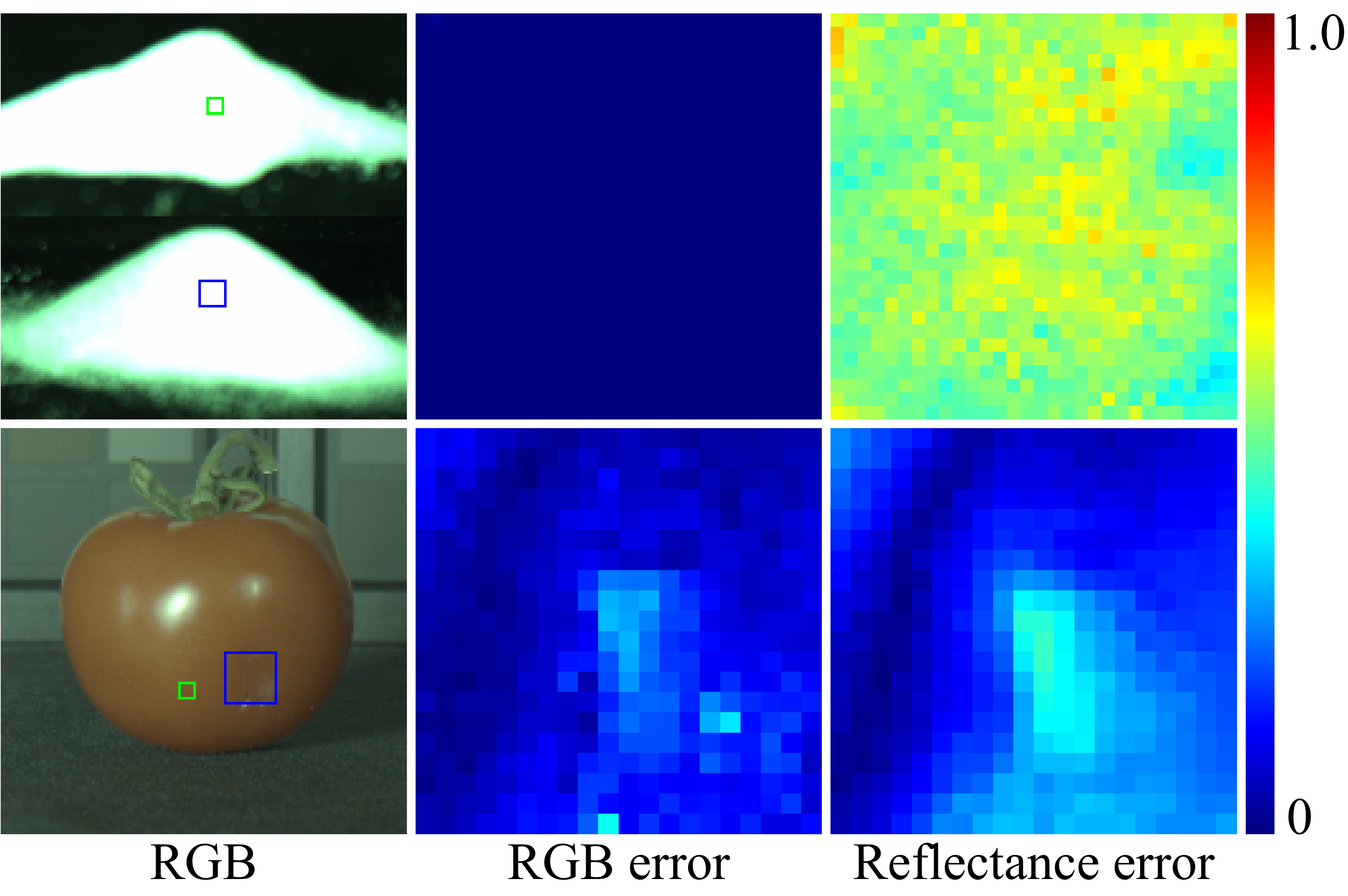}
\caption{The application results of recovered spectral reflectance. In each row, we randomly extract a pixel from the green box as the reference (the source material) and regard pixels from the blue box as the observation (the target material). A smaller green box is to reduce the variance of the reference. Then we calculate the error maps (MAE) between the reference and the observation for both RGB values and recovered spectral reflectances. The green and the blue box in the first row represent the salt and the sugar, respectively. The green and the blue box in the second row represent the flawless tomato peel and the region with a puncture, respectively. }
\label{fig:application}
\end{figure}

%% file: figures/limitation.tex
\begin{figure}
\centering
\includegraphics[width=0.47\textwidth]{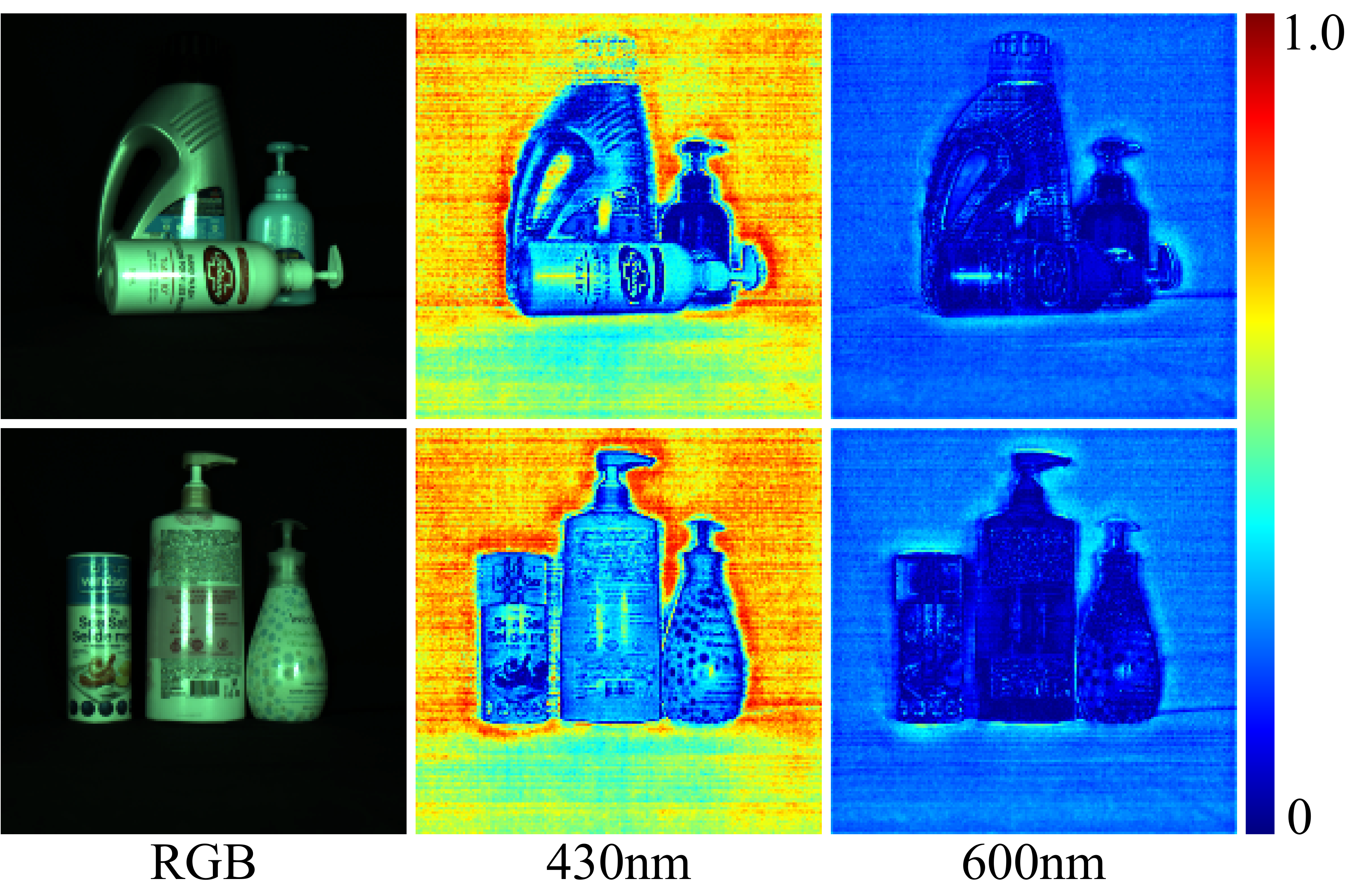}
\caption{Error maps of our recovered 430nm and 600nm bands.}
\label{fig:limitation}
\end{figure}

%% file: 5_conclusion.tex
\section{Conclusion}
This paper proposes a novel architecture motivated by the physical relationship between RGB images and the corresponding spectral reflectances, by which we estimate the components within the sub-space of the degradation matrix $\hat{\boldsymbol{\mathcal{H}}}$ to compensate for the final output. Our proposed architecture can be easily adapted to RGB images illuminated by more than one light source with only the output size of the auxiliary task needs to be changed. We also adopt meta-auxiliary learning to make use of the internal information of the input RGB images at test-time. Qualitative and quantitative evaluations demonstrate that our method surpasses state-of-the-art approaches by a large margin. Extensive ablation studies further justify the significant contribution of each component in our proposed method.

%% file: 6_acknowledgment.tex
\section{Acknowledgments}
\noindent The authors would like to thank the Natural Sciences and Engineering Research Council of Canada, the University of Alberta, and the University of Manitoba for the partial financial funding. 

%% file: figures/supp_qualitative_1.tex
\begin{figure*}[!htbp]
\centering
\includegraphics[width=0.98\textwidth]{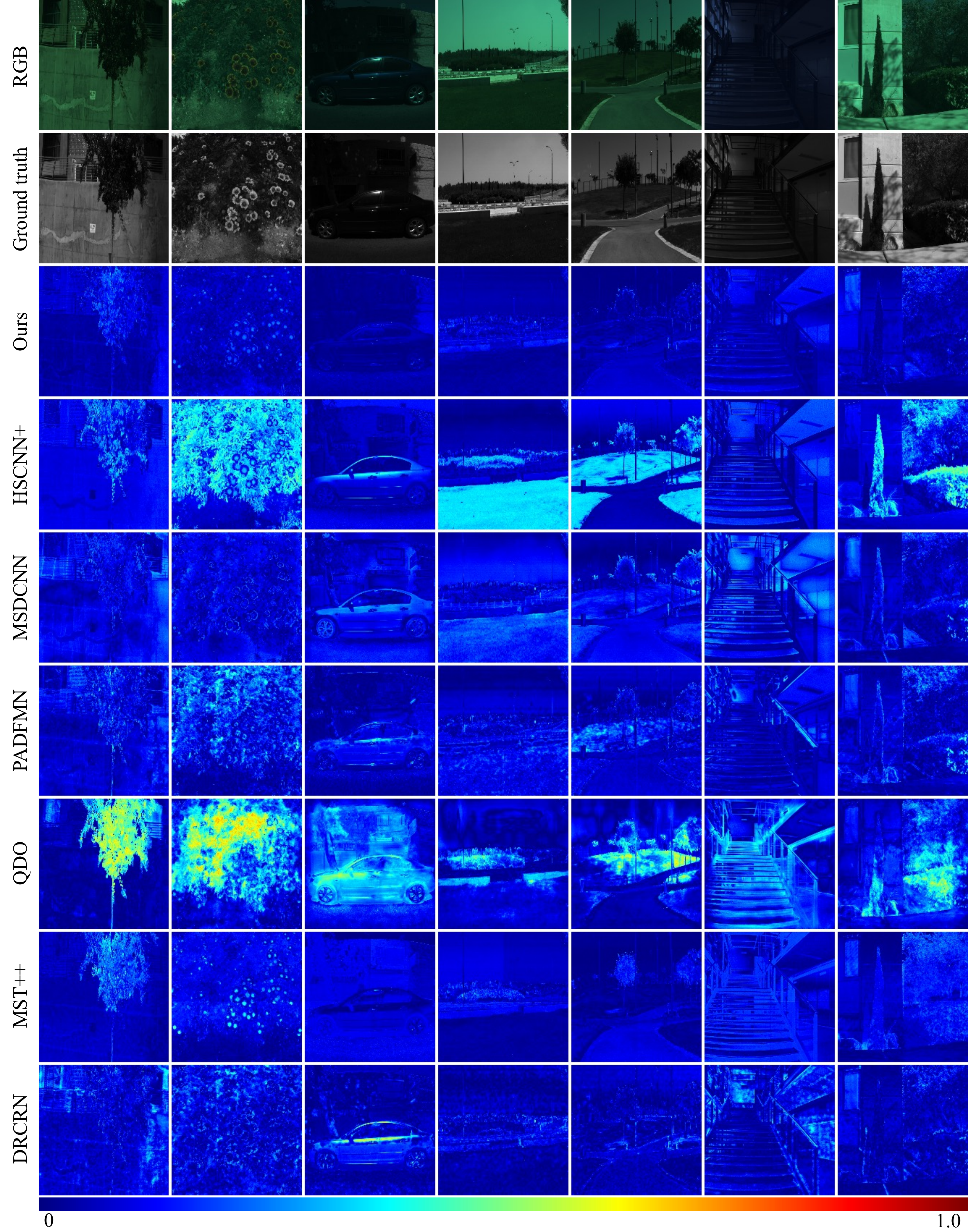}
\caption{More qualitative comparison of error maps (MAE between the recovered results and the ground truth) on synthetic data with state-of-the-art approaches. }
\label{fig:supp_qualitative_1}
\end{figure*}

%% file: figures/supp_qualitative_2.tex
\begin{figure*}[!htbp]
\centering
\includegraphics[width=0.98\textwidth]{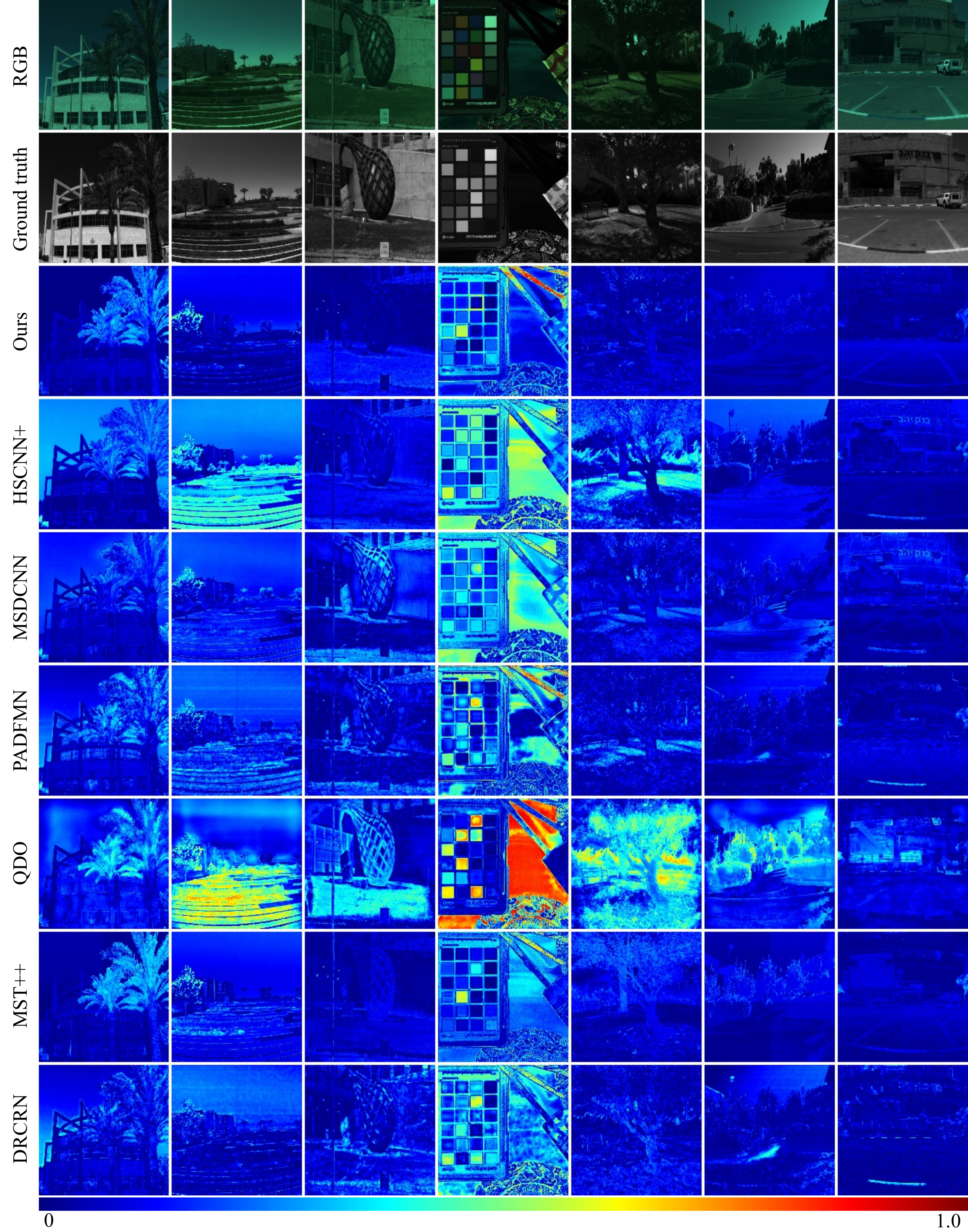}
\caption{More qualitative comparison of error maps (MAE between the recovered results and the ground truth) on synthetic data with state-of-the-art approaches. }
\label{fig:supp_qualitative_2}
\end{figure*}

%% file: figures/supp_qualitative_3.tex
\begin{figure*}[!htbp]
\centering
\includegraphics[width=0.98\textwidth]{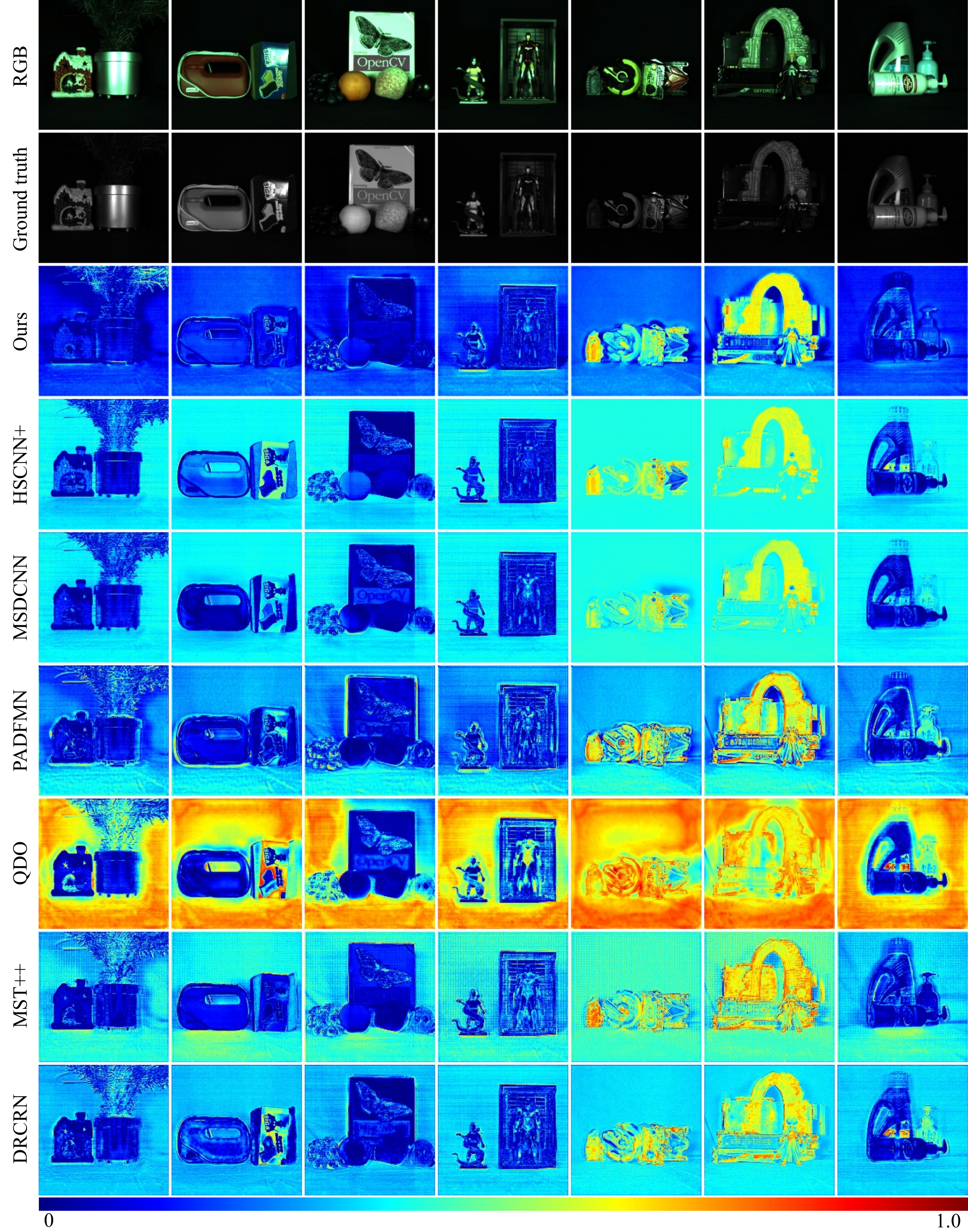}
\caption{More qualitative comparison of error maps (MAE between the recovered results and the ground truth) on real data with state-of-the-art approaches.}
\label{fig:supp_qualitative_3}
\end{figure*}

%% file: figures/synthetic_curve_1.tex
\begin{figure*}[!htbp]
\centering
\includegraphics[width=\textwidth]{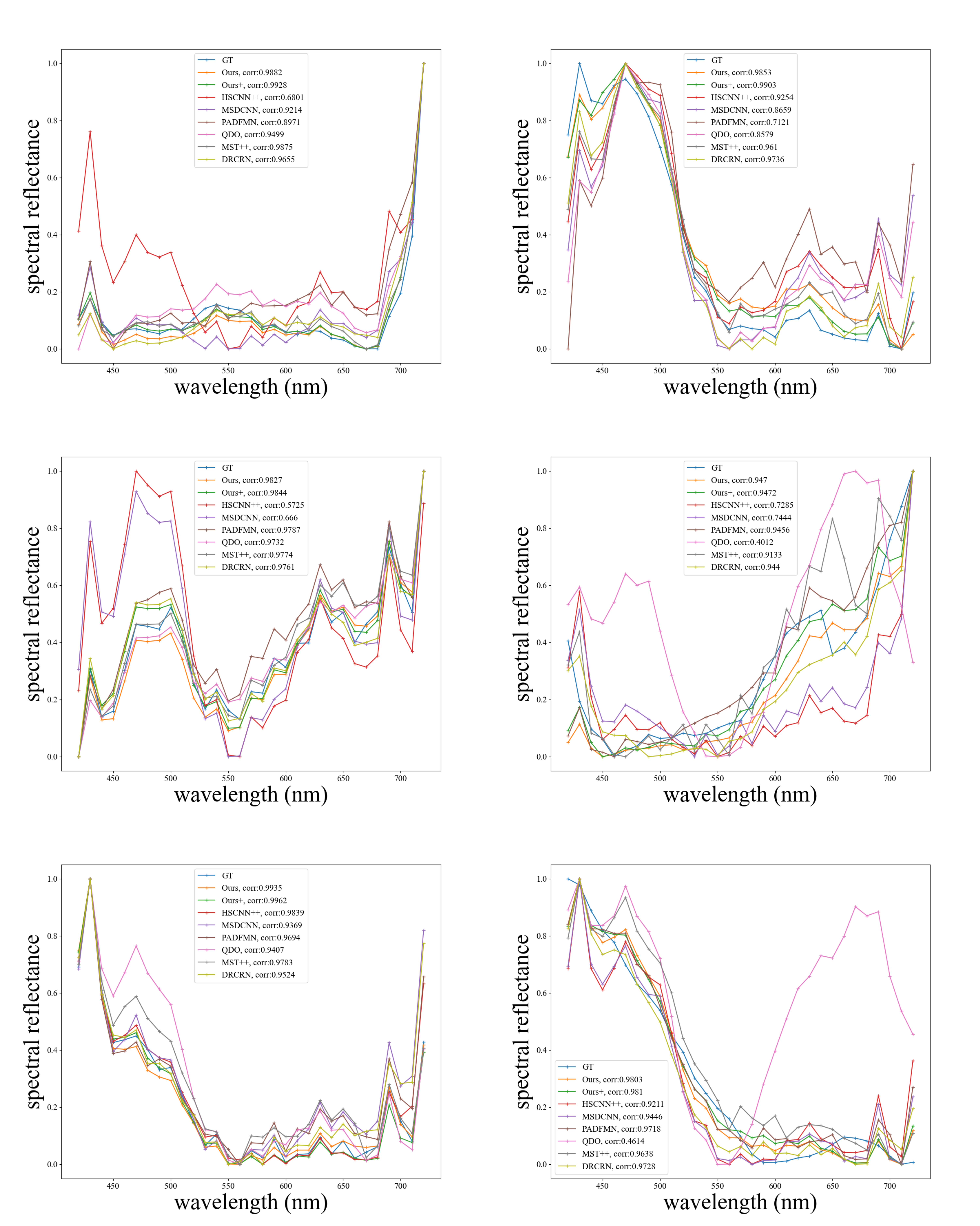}
\caption{Comparison of recovered spectral reflectance curves on synthetic data. We can see that our recovered spectral reflectance has higher correlation with the ground truth than that of other methods.}
\label{fig:synthetic_curve_1}
\end{figure*}

%% file: figures/synthetic_curve_2.tex
\begin{figure*}[!htbp]
\centering
\includegraphics[width=\textwidth]{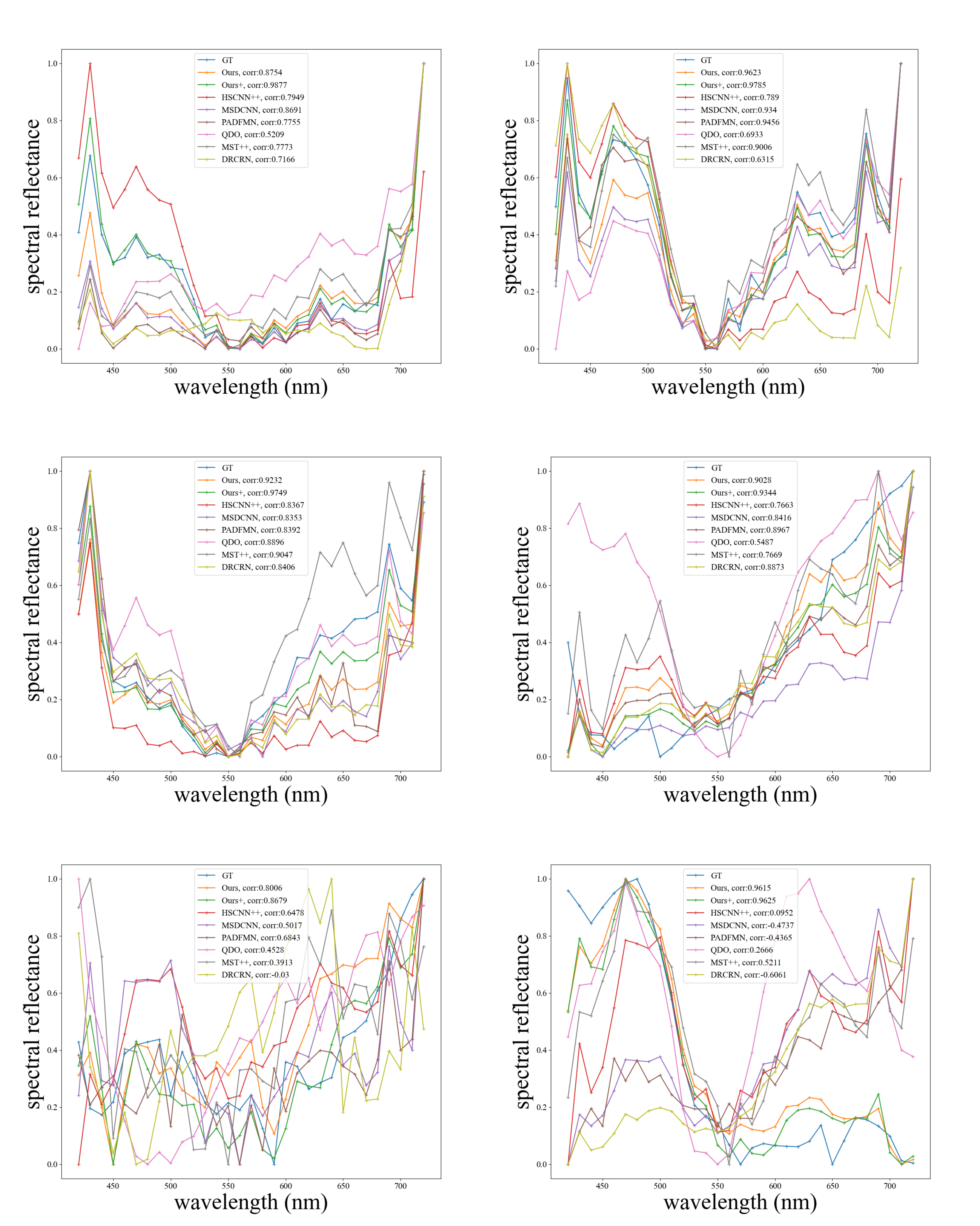}
\caption{Comparison of recovered spectral reflectance curves on synthetic data. We can see that when the quality of our recovered spectral reflectance under a single illumination is non-ideal, one more illumination can significantly improve the performance of our method.}
\label{fig:synthetic_curve_2}
\end{figure*}

%% file: figures/real_curve_1.tex
\begin{figure*}[!htbp]
\centering
\includegraphics[width=\textwidth]{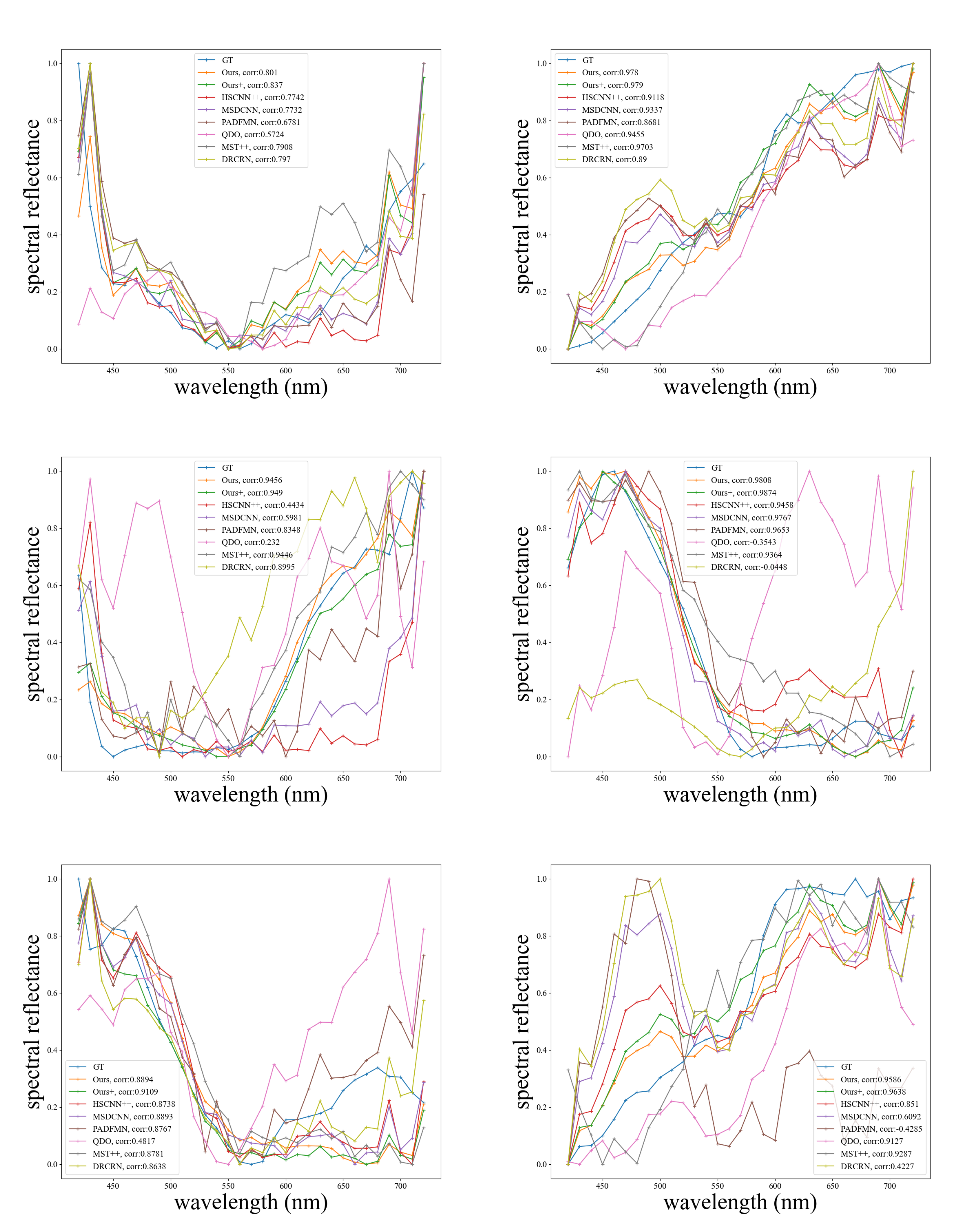}
\caption{Comparison of recovered spectral reflectance curves on real data. We can see that one more illumination can also help to improve the performance when testing on real data.}
\label{fig:real_curve_1}
\end{figure*}

%% file: figures/real_curve_2.tex
\begin{figure*}[!htbp]
\centering
\includegraphics[width=\textwidth]{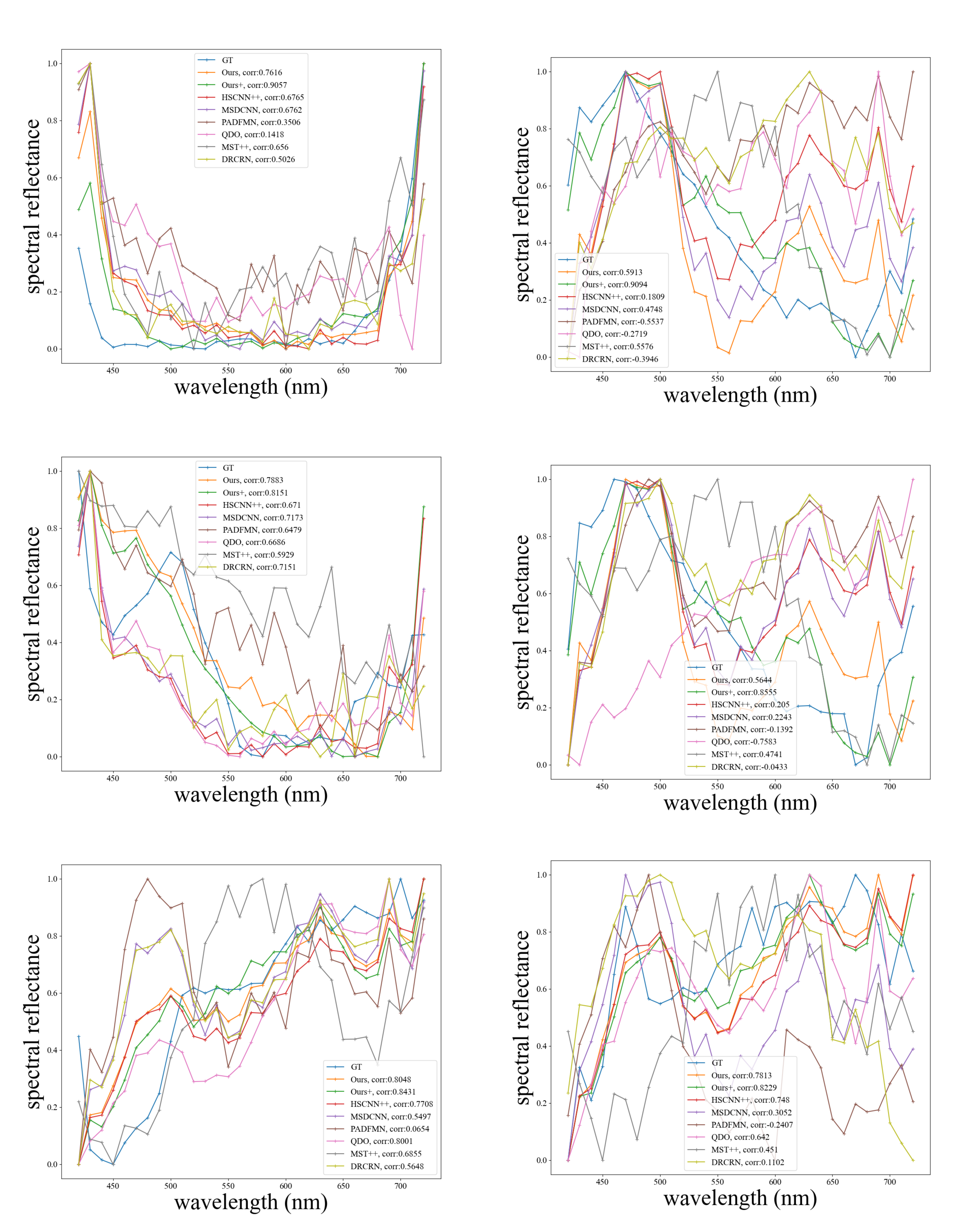}
\caption{Comparison of recovered spectral reflectance curves on real data. We can see that using a single illumination may still suffer from the domain gap and one more illumination can reduce this problem.}
\label{fig:real_curve_2}
\end{figure*}

%% file: figures/iphone.tex
\begin{figure*}[t]
\centering
\includegraphics[width=0.47\textwidth]{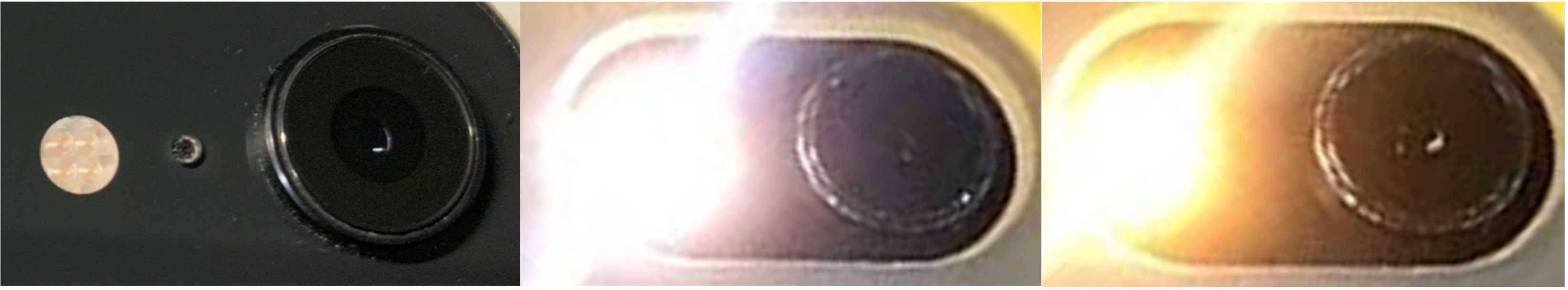}
\caption{True tone flash of an iPhone XR with LEDs off (left), white LEDs on (middle) and amber LEDs on (right). The middle and the right images are obtained from~\cite{youtube} which need jailbreak to change the color of flashlights.}
\label{fig:iphone}
\end{figure*}

%% file: main.bbl
\begin{thebibliography}{10}
\providecommand{\url}[1]{#1}
\csname url@samestyle\endcsname
\providecommand{\newblock}{\relax}
\providecommand{\bibinfo}[2]{#2}
\providecommand{\BIBentrySTDinterwordspacing}{\spaceskip=0pt\relax}
\providecommand{\BIBentryALTinterwordstretchfactor}{4}
\providecommand{\BIBentryALTinterwordspacing}{\spaceskip=\fontdimen2\font plus
\BIBentryALTinterwordstretchfactor\fontdimen3\font minus
  \fontdimen4\font\relax}
\providecommand{\BIBforeignlanguage}[2]{{%
\expandafter\ifx\csname l@#1\endcsname\relax
\typeout{** WARNING: IEEEtran.bst: No hyphenation pattern has been}%
\typeout{** loaded for the language `#1'. Using the pattern for}%
\typeout{** the default language instead.}%
\else
\language=\csname l@#1\endcsname
\fi
#2}}
\providecommand{\BIBdecl}{\relax}
\BIBdecl

\bibitem{goetz1985imaging}
A.~F. Goetz, G.~Vane, J.~E. Solomon, and B.~N. Rock, ``Imaging spectrometry for
  earth remote sensing,'' \emph{science}, vol. 228, no. 4704, pp. 1147--1153,
  1985.

\bibitem{nanni2006spectral}
M.~R. Nanni and J.~A.~M. Dematt{\^e}, ``Spectral reflectance methodology in
  comparison to traditional soil analysis,'' \emph{Soil science society of
  America journal}, vol.~70, no.~2, pp. 393--407, 2006.

\bibitem{carter1993responses}
G.~A. Carter, ``Responses of leaf spectral reflectance to plant stress,''
  \emph{American journal of botany}, vol.~80, no.~3, pp. 239--243, 1993.

\bibitem{martinez2005multispectral}
A.~Mart{\'\i}nez-Us{\'o}, F.~Pla, and P.~Garc{\'\i}a-Sevilla, ``Multispectral
  image segmentation by energy minimization for fruit quality estimation,'' in
  \emph{Pattern Recognition and Image Analysis: Second Iberian Conference,
  IbPRIA 2005, Estoril, Portugal, June 7-9, 2005, Proceedings, Part II
  2}.\hskip 1em plus 0.5em minus 0.4em\relax Springer, 2005, pp. 689--696.

\bibitem{tsumura2001medical}
N.~Tsumura, Y.~Miyake, and F.~H. Imai, ``Medical vision: measurement of skin
  absolute spectral-reflectance image and the application to component
  analysis,'' in \emph{Proceedings of the 3rd International Conference on
  Multispectral Color Science (MCS'01)}.\hskip 1em plus 0.5em minus 0.4em\relax
  Citeseer, 2001, pp. 25--28.

\bibitem{preece2002monte}
S.~J. Preece and E.~Claridge, ``Monte carlo modelling of the spectral
  reflectance of the human eye,'' \emph{Physics in Medicine \& Biology},
  vol.~47, no.~16, p. 2863, 2002.

\bibitem{kim2016smartphone}
S.~Kim, D.~Cho, J.~Kim, M.~Kim, S.~Youn, J.~E. Jang, M.~Je, D.~H. Lee, B.~Lee,
  D.~L. Farkas \emph{et~al.}, ``Smartphone-based multispectral imaging: system
  development and potential for mobile skin diagnosis,'' \emph{Biomedical
  optics express}, vol.~7, no.~12, pp. 5294--5307, 2016.

\bibitem{he2020hyperspectral}
Q.~He and R.~Wang, ``Hyperspectral imaging enabled by an unmodified smartphone
  for analyzing skin morphological features and monitoring hemodynamics,''
  \emph{Biomedical optics express}, vol.~11, no.~2, pp. 895--910, 2020.

\bibitem{wang2016recent}
N.-N. Wang, D.-W. Sun, Y.-C. Yang, H.~Pu, and Z.~Zhu, ``Recent advances in the
  application of hyperspectral imaging for evaluating fruit quality,''
  \emph{Food analytical methods}, vol.~9, no.~1, pp. 178--191, 2016.

\bibitem{elmasry2012meat}
G.~ElMasry, D.~F. Barbin, D.-W. Sun, and P.~Allen, ``Meat quality evaluation by
  hyperspectral imaging technique: an overview,'' \emph{Critical reviews in
  food science and nutrition}, vol.~52, no.~8, pp. 689--711, 2012.

\bibitem{hagen2013review}
N.~Hagen and M.~W. Kudenov, ``Review of snapshot spectral imaging
  technologies,'' \emph{Optical Engineering}, vol.~52, no.~9, pp.
  090\,901--090\,901, 2013.

\bibitem{elmasry2010principles}
G.~ElMasry and D.-W. Sun, ``Principles of hyperspectral imaging technology,''
  in \emph{Hyperspectral imaging for food quality analysis and control}.\hskip
  1em plus 0.5em minus 0.4em\relax Elsevier, 2010, pp. 3--43.

\bibitem{cao2017spectral}
B.~Cao, N.~Liao, and H.~Cheng, ``Spectral reflectance reconstruction from rgb
  images based on weighting smaller color difference group,'' \emph{Color
  Research \& Application}, vol.~42, no.~3, pp. 327--332, 2017.

\bibitem{wu2019reflectance}
G.~Wu, ``Reflectance spectra recovery from a single rgb image by adaptive
  compressive sensing,'' \emph{Laser Physics Letters}, vol.~16, no.~8, p.
  085208, 2019.

\bibitem{deeb2018spectral}
R.~Deeb, D.~Muselet, M.~Hebert, and A.~Tr{\'e}meau, ``Spectral reflectance
  estimation from one rgb image using self-interreflections in a concave
  object,'' \emph{Applied optics}, vol.~57, no.~17, pp. 4918--4929, 2018.

\bibitem{fu2018spectral}
Y.~Fu, Y.~Zheng, L.~Zhang, and H.~Huang, ``Spectral reflectance recovery from a
  single rgb image,'' \emph{IEEE Transactions on Computational Imaging},
  vol.~4, no.~3, pp. 382--394, 2018.

\bibitem{nguyen2014training}
R.~M. Nguyen, D.~K. Prasad, and M.~S. Brown, ``Training-based spectral
  reconstruction from a single rgb image,'' in \emph{Computer Vision--ECCV
  2014: 13th European Conference, Zurich, Switzerland, September 6-12, 2014,
  Proceedings, Part VII 13}.\hskip 1em plus 0.5em minus 0.4em\relax Springer,
  2014, pp. 186--201.

\bibitem{park2007multispectral}
J.-I. Park, M.-H. Lee, M.~D. Grossberg, and S.~K. Nayar, ``Multispectral
  imaging using multiplexed illumination,'' in \emph{2007 IEEE 11th
  International Conference on Computer Vision}.\hskip 1em plus 0.5em minus
  0.4em\relax IEEE, 2007, pp. 1--8.

\bibitem{pcmag_truetone}
P.~M. Encyclopedia, ``True tone flash,'' Retrieved from
  \url{https://www.pcmag.com/encyclopedia/term/true-tone-flash}, 2021.

\bibitem{sun2021tuning}
B.~Sun, J.~Yan, X.~Zhou, and Y.~Zheng, ``Tuning ir-cut filter for
  illumination-aware spectral reconstruction from rgb,'' in \emph{Proceedings
  of the IEEE/CVF Conference on Computer Vision and Pattern Recognition}, 2021,
  pp. 84--93.

\bibitem{cai2022mst++}
Y.~Cai, J.~Lin, Z.~Lin, H.~Wang, Y.~Zhang, H.~Pfister, R.~Timofte, and
  L.~Van~Gool, ``Mst++: Multi-stage spectral-wise transformer for efficient
  spectral reconstruction,'' in \emph{Proceedings of the IEEE/CVF Conference on
  Computer Vision and Pattern Recognition}, 2022, pp. 745--755.

\bibitem{alvarez2017adversarial}
A.~Alvarez-Gila, J.~Van De~Weijer, and E.~Garrote, ``Adversarial networks for
  spatial context-aware spectral image reconstruction from rgb,'' in
  \emph{Proceedings of the IEEE international conference on computer vision
  workshops}, 2017, pp. 480--490.

\bibitem{zhang2020pixel}
L.~Zhang, Z.~Lang, P.~Wang, W.~Wei, S.~Liao, L.~Shao, and Y.~Zhang,
  ``Pixel-aware deep function-mixture network for spectral super-resolution,''
  in \emph{Proceedings of the AAAI Conference on Artificial Intelligence},
  vol.~34, no.~07, 2020, pp. 12\,821--12\,828.

\bibitem{fu2018joint}
Y.~Fu, T.~Zhang, Y.~Zheng, D.~Zhang, and H.~Huang, ``Joint camera spectral
  sensitivity selection and hyperspectral image recovery,'' in
  \emph{Proceedings of the European Conference on Computer Vision (ECCV)},
  2018, pp. 788--804.

\bibitem{shi2018hscnn+}
Z.~Shi, C.~Chen, Z.~Xiong, D.~Liu, and F.~Wu, ``Hscnn+: Advanced cnn-based
  hyperspectral recovery from rgb images,'' in \emph{Proceedings of the IEEE
  Conference on Computer Vision and Pattern Recognition Workshops}, 2018, pp.
  939--947.

\bibitem{yan2018accurate}
Y.~Yan, L.~Zhang, J.~Li, W.~Wei, and Y.~Zhang, ``Accurate spectral
  super-resolution from single rgb image using multi-scale cnn,'' in
  \emph{Pattern Recognition and Computer Vision: First Chinese Conference, PRCV
  2018, Guangzhou, China, November 23-26, 2018, Proceedings, Part II 1}.\hskip
  1em plus 0.5em minus 0.4em\relax Springer, 2018, pp. 206--217.

\bibitem{park2020fast}
S.~Park, J.~Yoo, D.~Cho, J.~Kim, and T.~H. Kim, ``Fast adaptation to
  super-resolution networks via meta-learning,'' in \emph{Computer Vision--ECCV
  2020: 16th European Conference, Glasgow, UK, August 23--28, 2020,
  Proceedings, Part XXVII 16}.\hskip 1em plus 0.5em minus 0.4em\relax Springer,
  2020, pp. 754--769.

\bibitem{shocher2018zero}
A.~Shocher, N.~Cohen, and M.~Irani, ``“zero-shot” super-resolution using
  deep internal learning,'' in \emph{Proceedings of the IEEE conference on
  computer vision and pattern recognition}, 2018, pp. 3118--3126.

\bibitem{ren2020neural}
D.~Ren, K.~Zhang, Q.~Wang, Q.~Hu, and W.~Zuo, ``Neural blind deconvolution
  using deep priors,'' in \emph{Proceedings of the IEEE/CVF Conference on
  Computer Vision and Pattern Recognition}, 2020, pp. 3341--3350.

\bibitem{huo2022blind}
D.~Huo, A.~Masoumzadeh, R.~Kushol, and Y.-H. Yang, ``Blind image deconvolution
  using variational deep image prior,'' \emph{arXiv preprint arXiv:2202.00179},
  2022.

\bibitem{rasti2021undip}
B.~Rasti, B.~Koirala, P.~Scheunders, and P.~Ghamisi, ``Undip: Hyperspectral
  unmixing using deep image prior,'' \emph{IEEE Transactions on Geoscience and
  Remote Sensing}, vol.~60, pp. 1--15, 2021.

\bibitem{sidorov2019deep}
O.~Sidorov and J.~Yngve~Hardeberg, ``Deep hyperspectral prior: Single-image
  denoising, inpainting, super-resolution,'' in \emph{Proceedings of the
  IEEE/CVF International Conference on Computer Vision Workshops}, 2019, pp.
  0--0.

\bibitem{chi2021test}
Z.~Chi, Y.~Wang, Y.~Yu, and J.~Tang, ``Test-time fast adaptation for dynamic
  scene deblurring via meta-auxiliary learning,'' in \emph{Proceedings of the
  IEEE/CVF Conference on Computer Vision and Pattern Recognition}, 2021, pp.
  9137--9146.

\bibitem{lin2020physically}
Y.-T. Lin and G.~D. Finlayson, ``Physically plausible spectral reconstruction
  from rgb images,'' in \emph{Proceedings of the IEEE/CVF Conference on
  Computer Vision and Pattern Recognition Workshops}, 2020, pp. 532--533.

\bibitem{tschannerl2019hyperspectral}
J.~Tschannerl, J.~Ren, H.~Zhao, F.-J. Kao, S.~Marshall, and P.~Yuen,
  ``Hyperspectral image reconstruction using multi-colour and time-multiplexed
  led illumination,'' \emph{Optics and Lasers in Engineering}, vol. 121, pp.
  352--357, 2019.

\bibitem{goel2015hypercam}
M.~Goel, E.~Whitmire, A.~Mariakakis, T.~S. Saponas, N.~Joshi, D.~Morris,
  B.~Guenter, M.~Gavriliu, G.~Borriello, and S.~N. Patel, ``Hypercam:
  hyperspectral imaging for ubiquitous computing applications,'' in
  \emph{Proceedings of the 2015 ACM International Joint Conference on Pervasive
  and Ubiquitous Computing}, 2015, pp. 145--156.

\bibitem{chakrabarti2011statistics}
A.~Chakrabarti and T.~Zickler, ``Statistics of real-world hyperspectral
  images,'' in \emph{CVPR 2011}.\hskip 1em plus 0.5em minus 0.4em\relax IEEE,
  2011, pp. 193--200.

\bibitem{arad2016sparse}
B.~Arad and O.~Ben-Shahar, ``Sparse recovery of hyperspectral signal from
  natural rgb images,'' in \emph{Computer Vision--ECCV 2016: 14th European
  Conference, Amsterdam, The Netherlands, October 11--14, 2016, Proceedings,
  Part VII 14}.\hskip 1em plus 0.5em minus 0.4em\relax Springer, 2016, pp.
  19--34.

\bibitem{akhtar2018hyperspectral}
N.~Akhtar and A.~Mian, ``Hyperspectral recovery from rgb images using gaussian
  processes,'' \emph{IEEE transactions on pattern analysis and machine
  intelligence}, vol.~42, no.~1, pp. 100--113, 2018.

\bibitem{jia2017rgb}
Y.~Jia, Y.~Zheng, L.~Gu, A.~Subpa-Asa, A.~Lam, Y.~Sato, and I.~Sato, ``From rgb
  to spectrum for natural scenes via manifold-based mapping,'' in
  \emph{Proceedings of the IEEE international conference on computer vision},
  2017, pp. 4705--4713.

\bibitem{timofte2015a+}
R.~Timofte, V.~De~Smet, and L.~Van~Gool, ``A+: Adjusted anchored neighborhood
  regression for fast super-resolution,'' in \emph{Computer Vision--ACCV 2014:
  12th Asian Conference on Computer Vision, Singapore, Singapore, November 1-5,
  2014, Revised Selected Papers, Part IV 12}.\hskip 1em plus 0.5em minus
  0.4em\relax Springer, 2015, pp. 111--126.

\bibitem{xiong2017hscnn}
Z.~Xiong, Z.~Shi, H.~Li, L.~Wang, D.~Liu, and F.~Wu, ``Hscnn: Cnn-based
  hyperspectral image recovery from spectrally undersampled projections,'' in
  \emph{Proceedings of the IEEE International Conference on Computer Vision
  Workshops}, 2017, pp. 518--525.

\bibitem{zhao2020hierarchical}
Y.~Zhao, L.-M. Po, Q.~Yan, W.~Liu, and T.~Lin, ``Hierarchical regression
  network for spectral reconstruction from rgb images,'' in \emph{Proceedings
  of the IEEE/CVF Conference on Computer Vision and Pattern Recognition
  Workshops}, 2020, pp. 422--423.

\bibitem{stiebel2018reconstructing}
T.~Stiebel, S.~Koppers, P.~Seltsam, and D.~Merhof, ``Reconstructing spectral
  images from rgb-images using a convolutional neural network,'' in
  \emph{Proceedings of the IEEE Conference on Computer Vision and Pattern
  Recognition Workshops}, 2018, pp. 948--953.

\bibitem{li2022drcr}
J.~Li, S.~Du, C.~Wu, Y.~Leng, R.~Song, and Y.~Li, ``Drcr net: Dense residual
  channel re-calibration network with non-local purification for spectral super
  resolution,'' in \emph{Proceedings of the IEEE/CVF Conference on Computer
  Vision and Pattern Recognition}, 2022, pp. 1259--1268.

\bibitem{hang2021spectral}
R.~Hang, Q.~Liu, and Z.~Li, ``Spectral super-resolution network guided by
  intrinsic properties of hyperspectral imagery,'' \emph{IEEE Transactions on
  Image Processing}, vol.~30, pp. 7256--7265, 2021.

\bibitem{li2022quantization}
L.~Li, L.~Wang, W.~Song, L.~Zhang, Z.~Xiong, and H.~Huang, ``Quantization-aware
  deep optics for diffractive snapshot hyperspectral imaging,'' in
  \emph{Proceedings of the IEEE/CVF Conference on Computer Vision and Pattern
  Recognition}, 2022, pp. 19\,780--19\,789.

\bibitem{zhang2022implicit}
K.~Zhang, D.~Zhu, X.~Min, and G.~Zhai, ``Implicit neural representation
  learning for hyperspectral image super-resolution,'' \emph{IEEE Transactions
  on Geoscience and Remote Sensing}, vol.~61, pp. 1--12, 2022.

\bibitem{dong2021model}
W.~Dong, C.~Zhou, F.~Wu, J.~Wu, G.~Shi, and X.~Li, ``Model-guided deep
  hyperspectral image super-resolution,'' \emph{IEEE Transactions on Image
  Processing}, vol.~30, pp. 5754--5768, 2021.

\bibitem{wang2019deep}
W.~Wang, W.~Zeng, Y.~Huang, X.~Ding, and J.~Paisley, ``Deep blind hyperspectral
  image fusion,'' in \emph{Proceedings of the IEEE/CVF International Conference
  on Computer Vision}, 2019, pp. 4150--4159.

\bibitem{zhu2020hyperspectral}
Z.~Zhu, J.~Hou, J.~Chen, H.~Zeng, and J.~Zhou, ``Hyperspectral image
  super-resolution via deep progressive zero-centric residual learning,''
  \emph{IEEE Transactions on Image Processing}, vol.~30, pp. 1423--1438, 2020.

\bibitem{liu2019self}
S.~Liu, A.~Davison, and E.~Johns, ``Self-supervised generalisation with meta
  auxiliary learning,'' \emph{Advances in Neural Information Processing
  Systems}, vol.~32, 2019.

\bibitem{andrychowicz2016learning}
M.~Andrychowicz, M.~Denil, S.~Gomez, M.~W. Hoffman, D.~Pfau, T.~Schaul,
  B.~Shillingford, and N.~De~Freitas, ``Learning to learn by gradient descent
  by gradient descent,'' \emph{Advances in neural information processing
  systems}, vol.~29, 2016.

\bibitem{zhang2018fine}
Y.~Zhang, H.~Tang, and K.~Jia, ``Fine-grained visual categorization using
  meta-learning optimization with sample selection of auxiliary data,'' in
  \emph{Proceedings of the european conference on computer vision (ECCV)},
  2018, pp. 233--248.

\bibitem{finn2017model}
C.~Finn, P.~Abbeel, and S.~Levine, ``Model-agnostic meta-learning for fast
  adaptation of deep networks,'' in \emph{International conference on machine
  learning}.\hskip 1em plus 0.5em minus 0.4em\relax PMLR, 2017, pp. 1126--1135.

\bibitem{sun2020test}
Y.~Sun, X.~Wang, Z.~Liu, J.~Miller, A.~Efros, and M.~Hardt, ``Test-time
  training with self-supervision for generalization under distribution
  shifts,'' in \emph{International conference on machine learning}.\hskip 1em
  plus 0.5em minus 0.4em\relax PMLR, 2020, pp. 9229--9248.

\bibitem{sun2019meta}
Q.~Sun, Y.~Liu, T.-S. Chua, and B.~Schiele, ``Meta-transfer learning for
  few-shot learning,'' in \emph{Proceedings of the IEEE/CVF Conference on
  Computer Vision and Pattern Recognition}, 2019, pp. 403--412.

\bibitem{soh2020meta}
J.~W. Soh, S.~Cho, and N.~I. Cho, ``Meta-transfer learning for zero-shot
  super-resolution,'' in \emph{Proceedings of the IEEE/CVF Conference on
  Computer Vision and Pattern Recognition}, 2020, pp. 3516--3525.

\bibitem{wang2020deep}
Z.~Wang, J.~Chen, and S.~C. Hoi, ``Deep learning for image super-resolution: A
  survey,'' \emph{IEEE transactions on pattern analysis and machine
  intelligence}, vol.~43, no.~10, pp. 3365--3387, 2020.

\bibitem{guo2020closed}
Y.~Guo, J.~Chen, J.~Wang, Q.~Chen, J.~Cao, Z.~Deng, Y.~Xu, and M.~Tan,
  ``Closed-loop matters: Dual regression networks for single image
  super-resolution,'' in \emph{Proceedings of the IEEE/CVF conference on
  computer vision and pattern recognition}, 2020, pp. 5407--5416.

\bibitem{valada2018deep}
A.~Valada, N.~Radwan, and W.~Burgard, ``Deep auxiliary learning for visual
  localization and odometry,'' in \emph{2018 IEEE international conference on
  robotics and automation (ICRA)}.\hskip 1em plus 0.5em minus 0.4em\relax IEEE,
  2018, pp. 6939--6946.

\bibitem{lu2020depth}
K.~Lu, N.~Barnes, S.~Anwar, and L.~Zheng, ``From depth what can you see? depth
  completion via auxiliary image reconstruction,'' in \emph{Proceedings of the
  IEEE/CVF conference on computer vision and pattern recognition}, 2020, pp.
  11\,306--11\,315.

\bibitem{odena2017conditional}
A.~Odena, C.~Olah, and J.~Shlens, ``Conditional image synthesis with auxiliary
  classifier gans,'' in \emph{International conference on machine
  learning}.\hskip 1em plus 0.5em minus 0.4em\relax PMLR, 2017, pp. 2642--2651.

\bibitem{maas2013rectifier}
A.~L. Maas, A.~Y. Hannun, A.~Y. Ng \emph{et~al.}, ``Rectifier nonlinearities
  improve neural network acoustic models,'' in \emph{Proc. icml}, vol.~30,
  no.~1.\hskip 1em plus 0.5em minus 0.4em\relax Atlanta, Georgia, USA, 2013,
  p.~3.

\bibitem{dosovitskiy2015flownet}
A.~Dosovitskiy, P.~Fischer, E.~Ilg, P.~Hausser, C.~Hazirbas, V.~Golkov, P.~Van
  Der~Smagt, D.~Cremers, and T.~Brox, ``Flownet: Learning optical flow with
  convolutional networks,'' in \emph{Proceedings of the IEEE international
  conference on computer vision}, 2015, pp. 2758--2766.

\bibitem{sun2018pwc}
D.~Sun, X.~Yang, M.-Y. Liu, and J.~Kautz, ``Pwc-net: Cnns for optical flow
  using pyramid, warping, and cost volume,'' in \emph{Proceedings of the IEEE
  conference on computer vision and pattern recognition}, 2018, pp. 8934--8943.

\bibitem{tao2018scale}
X.~Tao, H.~Gao, X.~Shen, J.~Wang, and J.~Jia, ``Scale-recurrent network for
  deep image deblurring,'' in \emph{Proceedings of the IEEE conference on
  computer vision and pattern recognition}, 2018, pp. 8174--8182.

\bibitem{aiazzi2002context}
B.~Aiazzi, L.~Alparone, S.~Baronti, and A.~Garzelli, ``Context-driven fusion of
  high spatial and spectral resolution images based on oversampled
  multiresolution analysis,'' \emph{IEEE Transactions on geoscience and remote
  sensing}, vol.~40, no.~10, pp. 2300--2312, 2002.

\bibitem{monno2018single}
Y.~Monno, H.~Teranaka, K.~Yoshizaki, M.~Tanaka, and M.~Okutomi, ``Single-sensor
  rgb-nir imaging: High-quality system design and prototype implementation,''
  \emph{IEEE Sensors Journal}, vol.~19, no.~2, pp. 497--507, 2018.

\bibitem{lspdd}
J.~Roby and M.~Aubé, ``Lspdd: Lamp spectral power distribution database,''
  Retrieved from \url{https://lspdd.org/app/en/lamps/2629}, 2021.

\bibitem{jiang2013space}
J.~Jiang, D.~Liu, J.~Gu, and S.~S{\"u}sstrunk, ``What is the space of spectral
  sensitivity functions for digital color cameras?'' in \emph{2013 IEEE
  Workshop on Applications of Computer Vision (WACV)}.\hskip 1em plus 0.5em
  minus 0.4em\relax IEEE, 2013, pp. 168--179.

\bibitem{lowe1999object}
D.~G. Lowe, ``Object recognition from local scale-invariant features,'' in
  \emph{Proceedings of the seventh IEEE international conference on computer
  vision}, vol.~2.\hskip 1em plus 0.5em minus 0.4em\relax Ieee, 1999, pp.
  1150--1157.

\bibitem{dennison2004comparison}
P.~E. Dennison, K.~Q. Halligan, and D.~A. Roberts, ``A comparison of error
  metrics and constraints for multiple endmember spectral mixture analysis and
  spectral angle mapper,'' \emph{Remote Sensing of Environment}, vol.~93,
  no.~3, pp. 359--367, 2004.

\bibitem{korhonen2012peak}
J.~Korhonen and J.~You, ``Peak signal-to-noise ratio revisited: Is simple
  beautiful?'' in \emph{2012 Fourth International Workshop on Quality of
  Multimedia Experience}.\hskip 1em plus 0.5em minus 0.4em\relax IEEE, 2012,
  pp. 37--38.

\bibitem{wang2004image}
Z.~Wang, A.~C. Bovik, H.~R. Sheikh, and E.~P. Simoncelli, ``Image quality
  assessment: from error visibility to structural similarity,'' \emph{IEEE
  transactions on image processing}, vol.~13, no.~4, pp. 600--612, 2004.

\bibitem{kingma2014adam}
D.~P. Kingma and J.~Ba, ``Adam: A method for stochastic optimization,''
  \emph{arXiv preprint arXiv:1412.6980}, 2014.

\bibitem{loshchilov2016sgdr}
I.~Loshchilov and F.~Hutter, ``Sgdr: Stochastic gradient descent with warm
  restarts,'' \emph{arXiv preprint arXiv:1608.03983}, 2016.

\bibitem{li2019deep}
S.~Li, W.~Song, L.~Fang, Y.~Chen, P.~Ghamisi, and J.~A. Benediktsson, ``Deep
  learning for hyperspectral image classification: An overview,'' \emph{IEEE
  Transactions on Geoscience and Remote Sensing}, vol.~57, no.~9, pp.
  6690--6709, 2019.

\bibitem{bioucas2013hyperspectral}
J.~M. Bioucas-Dias, A.~Plaza, G.~Camps-Valls, P.~Scheunders, N.~Nasrabadi, and
  J.~Chanussot, ``Hyperspectral remote sensing data analysis and future
  challenges,'' \emph{IEEE Geoscience and remote sensing magazine}, vol.~1,
  no.~2, pp. 6--36, 2013.

\bibitem{thenkabail2013selection}
P.~S. Thenkabail, I.~Mariotto, M.~K. Gumma, E.~M. Middleton, D.~R. Landis, and
  K.~F. Huemmrich, ``Selection of hyperspectral narrowbands (hnbs) and
  composition of hyperspectral twoband vegetation indices (hvis) for
  biophysical characterization and discrimination of crop types using field
  reflectance and hyperion/eo-1 data,'' \emph{IEEE Journal of Selected Topics
  in Applied Earth Observations and Remote Sensing}, vol.~6, no.~2, pp.
  427--439, 2013.

\end{thebibliography}


\begin{thebibliography}{1}
\providecommand{\url}[1]{#1}
\csname url@samestyle\endcsname
\providecommand{\newblock}{\relax}
\providecommand{\bibinfo}[2]{#2}
\providecommand{\BIBentrySTDinterwordspacing}{\spaceskip=0pt\relax}
\providecommand{\BIBentryALTinterwordstretchfactor}{4}
\providecommand{\BIBentryALTinterwordspacing}{\spaceskip=\fontdimen2\font plus
\BIBentryALTinterwordstretchfactor\fontdimen3\font minus
  \fontdimen4\font\relax}
\providecommand{\BIBforeignlanguage}[2]{{%
\expandafter\ifx\csname l@#1\endcsname\relax
\typeout{** WARNING: IEEEtran.bst: No hyphenation pattern has been}%
\typeout{** loaded for the language `#1'. Using the pattern for}%
\typeout{** the default language instead.}%
\else
\language=\csname l@#1\endcsname
\fi
#2}}
\providecommand{\BIBdecl}{\relax}
\BIBdecl

\bibitem{henderson1981deriving}
H.~V. Henderson and S.~R. Searle, ``On deriving the inverse of a sum of
  matrices,'' \emph{Siam Review}, vol.~23, no.~1, pp. 53--60, 1981.

\bibitem{youtube}
iApplePro, ``How to change flashlight color on iphone xr/xs/12,'' Retrieved
  from \url{https://www.youtube.com/watch?v=jwkwtCZ_MrM}, 2021.

\bibitem{park2007multispectral}
J.-I. Park, M.-H. Lee, M.~D. Grossberg, and S.~K. Nayar, ``Multispectral
  imaging using multiplexed illumination,'' in \emph{2007 IEEE 11th
  International Conference on Computer Vision}.\hskip 1em plus 0.5em minus
  0.4em\relax IEEE, 2007, pp. 1--8.

\end{thebibliography}
